\documentclass{article}

\usepackage{microtype}
\usepackage{graphicx}
\usepackage{subcaption}
\usepackage{booktabs} 

\usepackage{hyperref}

\newcommand{\figllama}{share_12_dec_25/llama3-8B}
\newcommand{\figmistral}{share_12_dec_25/mistral-7B-v0.3}
\newcommand{\figqwen}{share_12_dec_25/qwen3-4B}

\usepackage[accepted]{icml2026}

\usepackage{amsmath}
\usepackage{amssymb}
\usepackage{mathtools}
\usepackage{amsthm}

\usepackage[capitalize,noabbrev]{cleveref}

\theoremstyle{plain}

\theoremstyle{definition}

\theoremstyle{remark}

\usepackage[textsize=tiny]{todonotes}

\icmltitlerunning{When Language Representations Interact: Separability and Cross-Lingual Effects in  LLMs}

\begin{document}

\twocolumn[
  \icmltitle{When Language Representations Interact: Separability and Cross-Lingual Effects in  LLMs}



  \icmlsetsymbol{equal}{*}

  \begin{icmlauthorlist}
    \icmlauthor{Boris Marinov}{equal,yyy}
    \icmlauthor{Angira Sharma}{equal,comp}
    \icmlauthor{Christian Schroeder de Witt}{comp}\\
    \icmlauthor{Philip Torr}{comp}
    \icmlauthor{Anisoara Calinescu}{comp}
    \icmlauthor{Jialin Yu}{comp}

  \end{icmlauthorlist}

  \icmlaffiliation{yyy}{UCL}
  \icmlaffiliation{comp}{University of Oxford}
  \icmlcorrespondingauthor{Boris Marinov}{boris.marinov.23@ucl.ac.uk}
  \icmlcorrespondingauthor{Angira Sharma}{angira.sharma@cs.ox.ac.uk}

  \icmlkeywords{Machine Learning, ICML}

  \vskip 0.3in
]



\printAffiliationsAndNotice{}  


\begin{abstract}
    Large language models exhibit strong multilingual capabilities, however, their internal representations are difficult to interpret.  Understanding these interactions is important for ensuring reliable behavior in multilingual systems. Recent work has shown that causal–geometric structure can explain how certain concepts are encoded as approximately linear and separable directions, but whether this framework extends to multilingual models, where language identity is correlated and hierarchical, is underexplored. We apply causal–geometric analysis to multilingual LLMs, studying 28 bilingual  contrasts across three models, allowing us to analyze when languages behave as approximately independent factors and when structured dependencies persist. We find evidence that language concepts admit stable linear representations that are largely separable under a covariance-adjusted (causal) inner product, with structured deviations reflecting linguistic similarity. Moreover, languages within the same family (such as Germanic or Romance) exhibit a simplex-like geometric structure, suggesting hierarchical organization. These results extend causal–geometric interpretability to multilingual settings and provide insight into how separability and similarity may exist  in multilingual LLM representations, motivating interpretability analyses that diagnose when and how structured dependencies between concepts can be anticipated. This has implications for trustworthy deployment, as residual structure between languages may lead to unintended cross-lingual effects when models are monitored or intervened upon.
\end{abstract}

\section{Introduction}

Large language models (LLMs) are increasingly deployed in multilingual settings, where predictable and reliable behavior across languages is critical for safety and control. Yet how languages are geometrically represented and related within these models remains underexplored. Language identity may correspond to a separable causal variable that can be independently monitored and intervened upon, or it may instead emerge from entangled linguistic features shaped by shared training data, scripts, and historical relatedness. This ambiguity poses a challenge for interpretability and steering methods, which typically assume that high-level concepts admit independent and targeted interventions \cite{arditi2024refusallanguagemodelsmediated, marks2024the, turner2024activationadditionsteeringlanguage}. In multilingual deployment, this question is not purely descriptive. Many
safety mechanisms (e.g., filtering, monitoring, or steering) are implemented
in one language and if language representations
are not well-separated, such interventions may produce unintended cross-lingual
effects, making behavior harder to predict and control.

A growing body of work suggests that many high-level concepts in neural networks admit approximately linear representations, enabling meaningful intervention-based analysis \citep{park2024linearrepresentationhypothesisgeometry, jiang2024originslinearrepresentationslarge}. Within this framework, concepts are often modeled as  causally independent to the extent that their associated directions in representation space are geometrically separable under a suitable inner product \cite{park2024geometrycategoricalhierarchicalconcepts}. 
Moreover, prior work has shown that multilingual representations exhibit rich geometric structure \citep{chang2022geometrymultilinguallanguagemodel, peng2022understandinglinearitycrosslingualword} showing that languages share structure.  However, similarity alone does not imply that language identity behaves as an independent, intervenable variable. In this work, we apply causal–geometric diagnostics to distinguish between similarity and approximate causal separability in multilingual LLMs. Moreover, it remains unclear whether this structure extends to language identity as a concept, particularly in the multilingual setting, where languages share lexical and orthographic features, reflect historical lineage, and are unevenly represented in training data. Taken together, these factors make multilinguality a useful setting for examining causal–geometric interpretability, particularly whether linear separability and independent intervention extend to correlated and hierarchical concepts.

We study whether language identity in LLMs behaves as an approximately causally separable variable, rather than merely a correlated feature. Concretely, we address three questions: First, can bilingual language differences be represented by stable directions in representation space? Second, are these directions approximately independent across languages, or does linguistic similarity induce structured overlap? Third, how does higher-level organization, including language families, emerge geometrically within LLM representations?

To address these questions, we analyze multilingual representations using a geometry-based interpretability framework \cite{park2024linearrepresentationhypothesisgeometry, park2024geometrycategoricalhierarchicalconcepts}. We adapt a previously proposed causal inner product and study 28 bilingual
language contrasts (plus 3 non-linguistic concepts) across three public LLMs (Qwen3-4B, Mistral-7B-v0.3, Llama-3-8B). Intuitively, this inner product rescales representation space such that directions corresponding to independent concepts become approximately orthogonal, allowing geometric overlap to be interpreted as residual dependence rather than arbitrary scaling. 


\textbf{Contributions:}
\begin{enumerate}
    \item We show that many bilingual contrasts admit stable, approximately
linear directions that generalize beyond estimation pairs, enabling
interpretable diagnostics of language identity.

\item We find that a covariance-adjusted inner product reduces overlap
between language directions, allowing residual structure to be interpreted
as meaningful similarity rather than noise.

\item We show that languages within the same family exhibit simplex-like
organization, revealing a geometric signature of hierarchical structure.

\item We use additive interventions as a diagnostic, finding systematic
logit-level effects but limited generation-level control, suggesting 
limits of linear steering methods in multilingual settings.
\end{enumerate}

At a high level, we find that languages are largely separable but not independent:
residual structure aligns with linguistic similarity and can explain when
cross-lingual effects are likely to occur. Together, these results indicate that multilingual representations in LLMs
combine approximate causal separability, hierarchical organization, and graded
linguistic similarity. This is relevant for multilingual safety and deployment, as residual structure may shape how effects propagate across languages even when representations are approximately separable.

\section{Related Work}

Our work builds on a growing body of research that studies the internal geometry and interpretability of LLMs. Early work on representation structure identified linear regularities in word embeddings and analogy spaces, suggesting that semantic and syntactic relations correspond to approximately linear directions \citep{mikolov-etal-2013-linguistic, gladkova-etal-2016-analogy, fournier-etal-2020-analogies}. Subsequent work extended these ideas to contextual and transformer-based models, showing that relational and categorical information can often be decoded linearly from hidden representations \citep{chiang-etal-2020-understanding, hernandez2024linearityrelationdecodingtransformer}.

More recent interpretability work has formalized these observations through the \emph{linear representation hypothesis}, which models high-level concepts as causal variables associated with stable linear directions in representation space \citep{park2024linearrepresentationhypothesisgeometry, jiang2024originslinearrepresentationslarge}. This framework has been used to analyze concept separability, partial orthogonality, and hierarchical structure in LLMs  \citep{jiang2023uncoveringmeaningsembeddingspartial, park2024geometrycategoricalhierarchicalconcepts}. Related work has also explored activation-based steering and concept control via linear interventions \citep{turner2024activationadditionsteeringlanguage, wang2024conceptalgebrascorebasedtextcontrolled}, while highlighting limitations of purely linear control  \citep{engels2024languagemodelfeatureslinear}.

Multilingual representation structure has primarily been studied from a geometric or probing perspective, showing that languages often occupy structured subspaces and exhibit parallelism across models \citep{chang2022geometrymultilinguallanguagemodel, peng2022understandinglinearitycrosslingualword, singh-etal-2019-bert, choenni2020does}.  However, the relationship between causal separability, linguistic similarity, and hierarchical organization in multilingual representations has not been systematically analyzed.

Our work bridges these lines of research by applying causal–geometric diagnostics to language concepts. We provide empirical evidence that bilingual  contrasts admit approximately linear, causally separable representations, while language families form hierarchical simplex structures with graded similarity, offering a unified view of multilingual geometry and causal separation in LLMs.

\section{Methodology}

Our analysis builds on the causal–geometric framework introduced in \cite{park2024linearrepresentationhypothesisgeometry, park2024geometrycategoricalhierarchicalconcepts} (for full theoretical development and proofs, we refer the reader to these works). To make the paper self-contained, we restate the core concepts and notation, with minor adaptations required for multilingual representations.

We adopt the causal framework for concept representations introduced in prior work \cite{park2024linearrepresentationhypothesisgeometry, park2024geometrycategoricalhierarchicalconcepts},
in which a concept $C$ is treated as a causal variable whose value can be intervened on
while holding other factors fixed. A language concept refers to a variable indicating the language (e.g., French), and a bilingual contrast refers to a pairwise difference (e.g., English→French). When a concept admits a linear representation,
counterfactual pairs that differ only in $C$ induce approximately parallel difference
vectors in representation space. Two concepts $C_1$ and $C_2$ are said to be \emph{causally
separable} if intervening on one does not affect the distribution of the other. Under an
appropriate inner product, causally separable concepts correspond to approximately
orthogonal directions.  

To make these notions operational, we require an inner product on representation space
that is consistent with causal separability. We  discuss  the causal inner product \cite{park2024linearrepresentationhypothesisgeometry},
which provides a unified geometric framework for analyzing embedding and unembedding
representations.

    \subsection{Causal Inner Product}

We work with two representation spaces: the embedding (context) space, $\lambda$,
and the unembedding (output) space, $\gamma$. A concept $W$ admits representations
$\bar{\lambda}_W$ and $\bar{\gamma}_W$. For any concept $Z$ that is causally
separable from $W$,
\begin{equation}
\bar{\lambda}_W^\top \bar{\gamma}_W > 0,
\quad
\bar{\lambda}_W^\top \bar{\gamma}_Z = 0
\label{eq:causal_separability}
\end{equation}

The causal inner product  aligns embedding and
unembedding representations such that causally separable concepts are orthogonal.
Let $\Sigma_\gamma = \mathrm{Cov}[\gamma(y)]$ denote the covariance of unembedding
vectors over the vocabulary. 

The causal inner product provides a geometry in which independent concepts tend to correspond to approximately orthogonal directions. In standard Euclidean space, correlations in the data can distort angles, making this relationship less clear. The causal inner product mitigates this effect by accounting for covariance in the representation space. The causal inner product is defined as
\begin{equation}
\langle \gamma, \gamma' \rangle_{\mathbb{C}}
= \gamma^\top \Sigma_\gamma^{-1} \gamma'
\label{eq:causal_inner_product}
\end{equation}

 where $\gamma’$ denotes an arbitrary vector in unembedding space.

 For each pair of concepts $(C,Z)$, we compute the causal inner product
\begin{equation}
    M_{CZ} = \langle \gamma_C, \gamma_Z \rangle_{\mathbb{C}}
\end{equation}

using the covariance-adjusted inner product defined in \eqref{eq:causal_inner_product}.
We analyze the resulting matrix to assess causal separability:
near-zero off-diagonal entries ( $\langle \gamma_{C_i}, \gamma_{C_j} \rangle \approx 0$ ) indicate approximate separability,
while structured deviations reflect residual association between concepts.

An equivalent formulation is obtained via the causal whitening transform $
g(y) = \Sigma_\gamma^{-1/2} \gamma(y)$
under which the Euclidean inner product corresponds to the causal inner product,
i.e., $\langle \bar{\gamma}_W, \bar{\gamma}_Z \rangle_{\mathbb{C}}
= \bar{g}_W^\top \bar{g}_Z$.


\subsection{Concept Estimation}

For each of the concepts $W$, we estimate a mean unembedding direction
$\bar{\gamma}_W$ by averaging over all counterfactual word pairs and normalizing
with respect to the causal inner product,
\begin{equation}
\bar{\gamma}_W :=
\frac{\tilde{\gamma}_W}{\sqrt{\langle \tilde{\gamma}_W, \tilde{\gamma}_W \rangle_{\mathbb{C}}}},
\quad
\\
\text{where} 
\quad
\tilde{\gamma}_W =
\frac{1}{n_W} \sum_{i=1}^{n_W}
\big(\gamma(y_i(1)) - \gamma(y_i(0))\big)
\label{eq:gamma}
\end{equation}

\paragraph{Linear Measurement of Contexts}
To assess whether concept directions generalize beyond counterfactual word
pairs, we project embedding representations $\lambda(x)$ of natural language
contexts onto learned concept directions $\gamma_C$.
We focus on embedding and unembedding spaces because they directly mediate input representation and output prediction, providing a tractable view of features that influence model behavior.

\subsection{Hierarchy}

To analyze hierarchical structure, we transform unembedding vectors using the
causal whitening transform
\begin{equation}
\mathbf{g}(y) = \mathrm{Cov}(\gamma)^{-1/2}\big(\gamma(y) - \mathbb{E}[\gamma]\big)
\label{eq:whiten}
\end{equation}
which aligns embedding and unembedding representations under the causal inner
product. Centering by $\mathbb{E}[\gamma]$ ensures that tokens not associated
with a given concept have approximately zero projection in the transformed
space.

The causal whitening transform places all concept directions in a common Euclidean
space where distances and angles reflect the causal inner product. We then treat
each language-specific direction as a point in this space. For languages within the same family, we analyze whether their corresponding
points lie in a shared low-dimensional subspace.

Our empirical analysis proceeds in five steps: estimating bilingual concept
directions, measuring separability under a covariance-adjusted inner product,
testing whether directions separate natural-language contexts, analyzing
language-family geometry, and using additive interventions as a diagnostic of
logit-level effects.
\section{Experimental Design}

\paragraph{Languages and Data}
We study eight languages drawn from two families: Germanic (English, German, Dutch, Swedish) and Romance (French, Spanish, Italian, Portuguese), following standard linguistic classifications \cite{HEERINGA2023103512}. We select the four most widely spoken languages from each family to increase the likelihood of adequate representation in LLM training data. 

For hierarchical analyses, we treat Germanic and Romance as the only higher-level categories, without modeling sub-family structure. We construct language-family sets $Y(\text{Germanic})$ and $Y(\text{Romance})$, along with their constituent languages, and analyze the geometry of their representations in the transformed space.  We use two families to provide a controlled hierarchical setting rather than maximize coverage.

Token pairs for English, German, French, and Spanish are obtained from the word2word bilingual lexicon \cite{choe2019word2wordcollectionbilinguallexicons}. For Dutch, Swedish, Portuguese, and Italian, translations are generated using Google Translate. All candidate pairs are filtered to retain only single-token entries in the target model vocabularies. Using 3 semantic relations, we construct 28 bilingual  contrasts (plus 3 non-linguistic concepts), yielding 31 single-token counterfactual sets.  Contexts are sampled from bilingual Wikipedia corpora, and projections are analyzed by comparing the resulting distributions across languages.

In our multilingual setting, bilingual word translations are used as approximate counterfactual pairs, and should be viewed as a practical proxy for ideal interventions rather than ground-truth causal manipulations.   Despite this limitation, the consistency of results across multiple contrasts and models suggests that the observed structure is not driven solely by lexical artifacts.
\paragraph{Models} 
We evaluate on three  LLMs, Llama-3-8B, Mistral-7B-v0.3 and Qwen3-4B. We compute concept directions separately for each model and use the same pipeline for all hierarchy analyses.


\section{Results}

For brevity, we discuss the results for Qwen3-4B here, the results for other models are available in the Appendix.

\subsection{Linear Directions}
\label{results:linear_directions}
\begin{figure}[h]
  \centering
  \includegraphics[width=0.9\columnwidth]{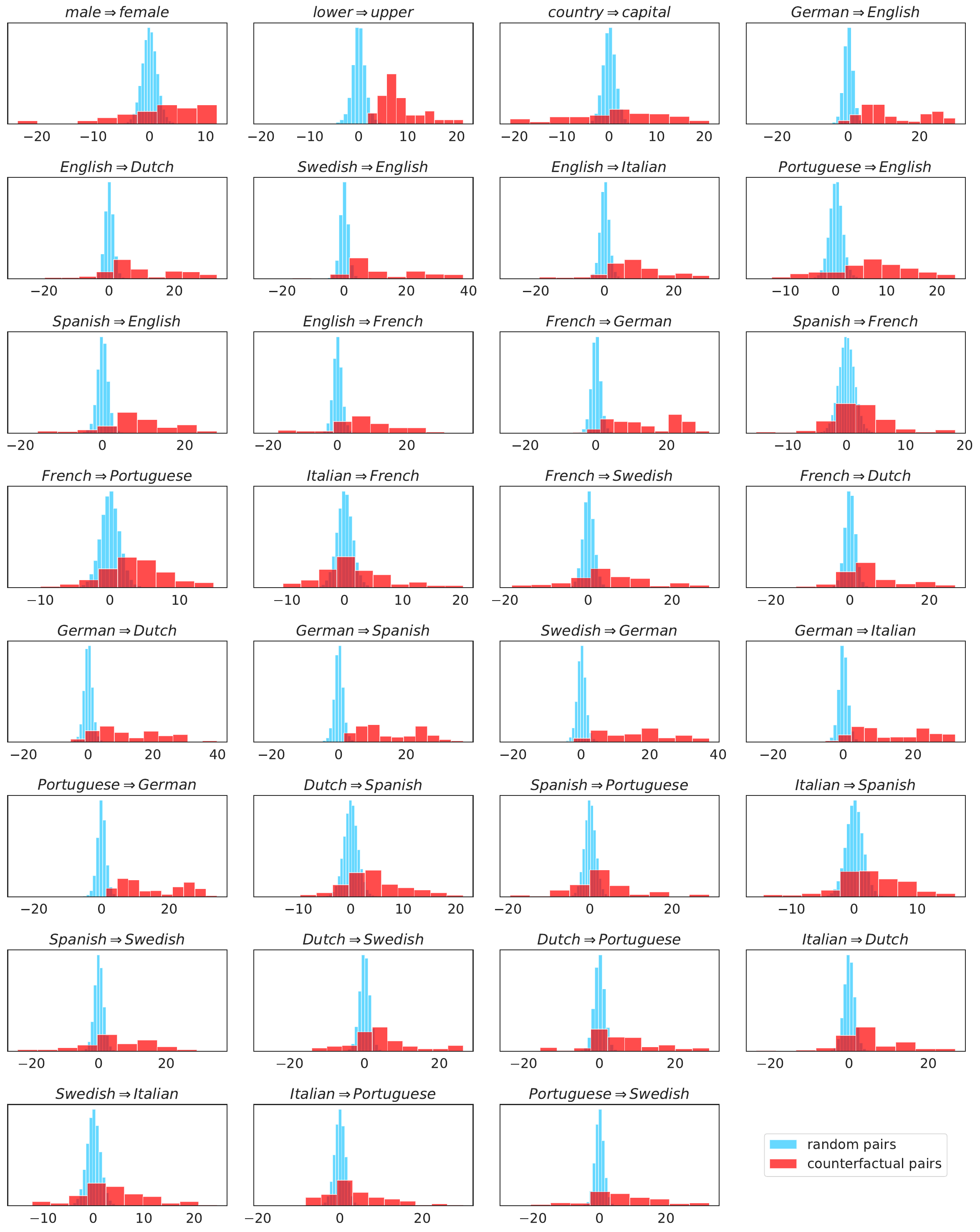}
  \caption{Projection histograms for Qwen3-4B for concepts, including
non-linguistic and bilingual  contrasts. Counterfactual word
differences (red) are projected onto the corresponding concept direction
$\bar{\gamma}_W$ using the causal inner product and compared against baseline of random
word differences (blue). Projections are computed using a leave-one-out
estimate $\bar{\gamma}_{W,(-i)}$ to avoid bias. The consistent right-skew of counterfactual projections, relative to random baselines centered near zero, indicates the presence of stable, approximately linear concept representations.
}
\label{fig: linear_directions_qwen}
\end{figure}

We first test whether bilingual contrasts admit stable linear representations. Figure~\ref{fig: linear_directions_qwen} shows projections of counterfactual word
differences onto their corresponding concept directions, compared against a baseline of 
 random word differences. Random differences have mean projection near
zero, whereas counterfactual pairs exhibit a pronounced right-skewed
distribution, indicating association with the estimated concept direction.

Across all concepts, we observe three regimes of behavior: clean separation
for simple categorical concepts (e.g., \textit{male}$\Rightarrow$\textit{female}),
partial separation for most bilingual  contrasts, and degraded or
entangled representations for closely related or underrepresented languages. Across bilingual contrasts, most language pairs exhibit clear association with
their estimated concept directions. For example, English--French and
German--English show pronounced right-skewed projection distributions, whereas
closely related pairs such as Dutch--Swedish exhibit weaker but still
non-random association. 

These results suggest stable linear signal across a range of language contrasts, motivating subsequent analysis of separability and hierarchical structure.

\begin{figure}[h]
  \centering
  \includegraphics[width=0.9\columnwidth]{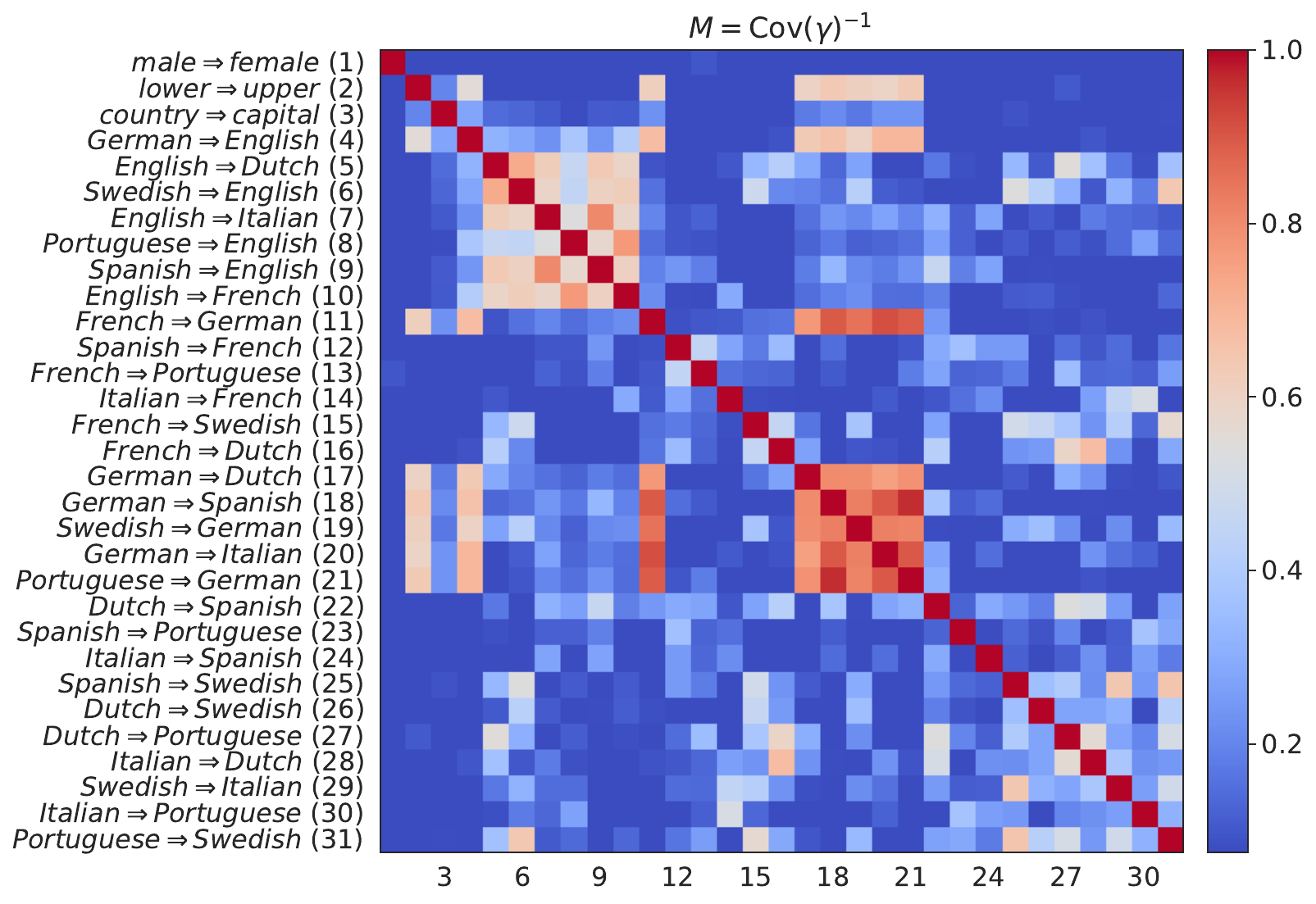}
  \caption{Causal inner product in Qwen-3-4B. Causally separable concepts are approximately orthogonal under the estimated causal inner product $\langle \bar{\gamma}, \bar{\gamma}’ \rangle_\mathbb{C} = \bar{\gamma}^\top \mathrm{Cov}(\gamma)^{-1}\bar{\gamma}’$. The heatmap shows $\lvert\langle \bar{\gamma}_W, \bar{\gamma}_Z\rangle\rvert$ across 31 concept pairs, with near-zero values for separable concepts and larger values for causally related or poorly represented languages.}
  \label{fig:qwen_heatmap}
\end{figure}


\begin{figure}[h]
  \centering
  \includegraphics[width=0.9\columnwidth]{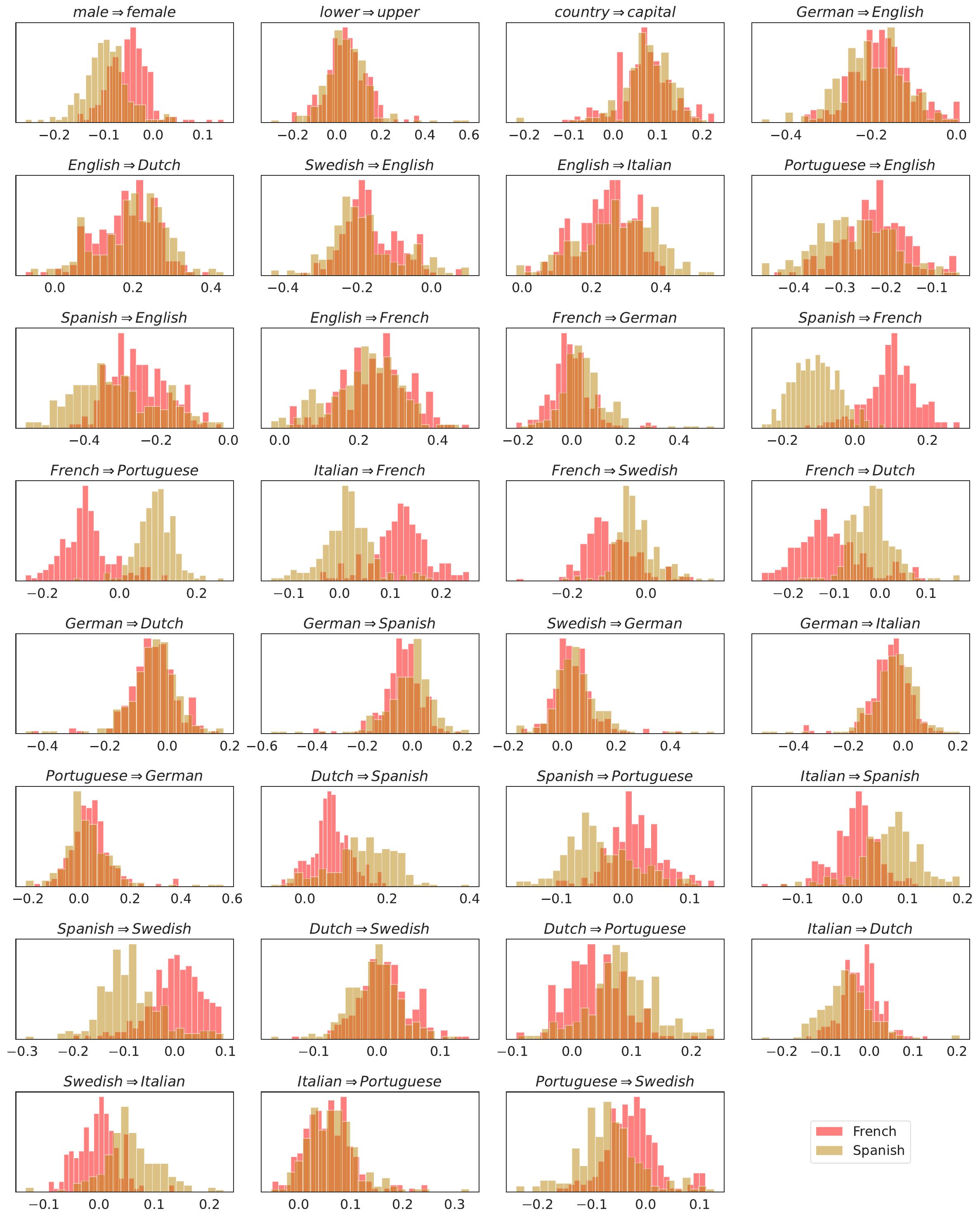}
  \caption{Histogram of $\mathbf{\gamma}_C^\top \lambda(x_j^{\text{fr}})$ vs $\mathbf{\gamma}_C^\top \lambda(x_j^{\text{es}})$ for  concepts $C$  for Qwen3-4B, where $\{ x_j^{\text{fr}} \}$ are random contexts from French Wikipedia, and $\{ x_j^{\text{es}} \}$ are random contexts from Spanish Wikipedia. Contexts from French
($x_j^{\text{fr}}$) and Spanish ($x_j^{\text{es}}$) exhibit clear separation
along concept directions involving either French or Spanish, while remaining
overlapping for unrelated concepts.
}
\label{fig: hist_classification_qwen}
\end{figure}

 \begin{figure*}[t]
  \centering
  \includegraphics[width=\textwidth]{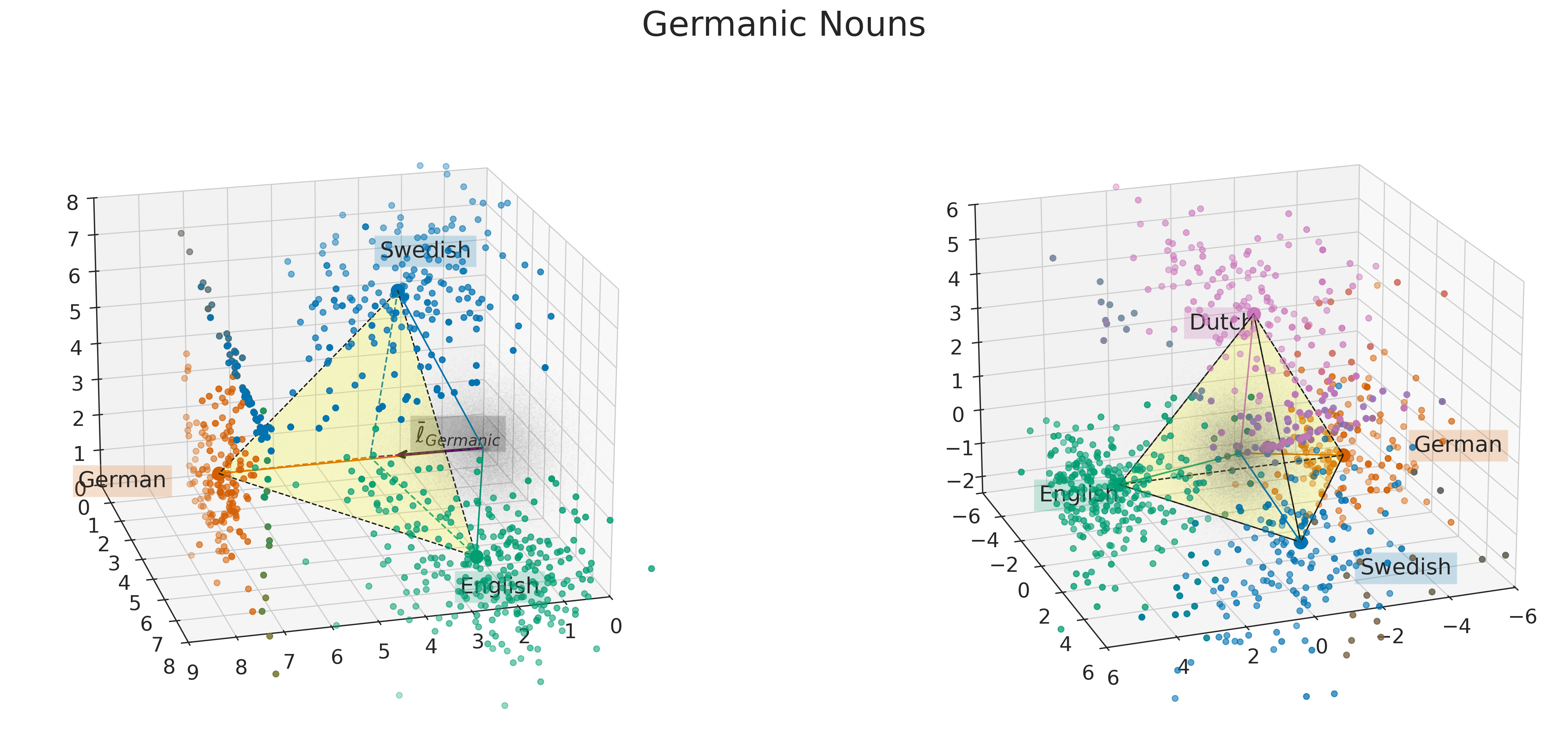}
  \caption{Qwen3-4B Languages within the Germanic family exhibit a simplex-like geometric structure 
  in representation space. Each vertex corresponds to a language-specific
  concept direction, while the simplex as a whole is approximately orthogonal
  to the Germanic family direction under the causal inner product.}
  \label{fig:germanic_simplex_qwen}
\end{figure*}

\subsection{Causal Separability}
\label{results:causal_separability}
We next test whether these representations are approximately separable under the causal inner product. Figure~\ref{fig:qwen_heatmap} shows that causal inner product heatmaps exhibit
near-orthogonality between most bilingual concept pairs that do not share a
language.  While many concept pairs appear approximately orthogonal, this can arise in high-dimensional spaces after whitening. We therefore focus on which pairs deviate from orthogonality, and find that these deviations align with known linguistic relationships, suggesting that the observed structure reflects non-random relationships. Importantly, the representations are near, but not exactly, orthogonal; overlap is not uniform across concepts, but instead aligns with known linguistic relationships, indicating that deviations from orthogonality are more consistent with linguistic similarity than with random noise. This distinction is relevant for safety and reliability analysis, as structured overlap may help anticipate some cross-concept effects.

\subsection{Classification and Linguistic Similarity}
\label{results:classification_linguistic_similarity}
Fig.~\ref{fig: hist_classification_qwen} shows the histograms of $\gamma_C^\top \lambda(x)$ for bilingual contexts which exhibits clear
separation along language-related concept directions, while remaining centered
near zero for unrelated concepts. As illustrated by the
Spanish--Portuguese case, linguistically similar but causally separable
language pairs exhibit weaker but non-trivial linear separation. 

For French and Spanish contexts, concept directions involving either French or
Spanish produce strong distributional separation, whereas off-target concepts
such as \textit{male}$\Rightarrow$\textit{female} or \textit{German}$\Rightarrow$\textit{English}
show substantial overlap. Notably, concept directions associated with
linguistically related but causally separable languages (e.g.,
\textit{Italian}$\Rightarrow$\textit{Portuguese}) also exhibit non-trivial
separation, reflecting shared linguistic structure under the causal inner
product.

 Appendix \ref{app:steering} shows that additive interventions induce  log-odds shifts across many concepts, providing evidence of representation-level causal signal even when generation-level behavior remains limited. This further  suggests that the identified structure is not purely descriptive, but reflects features that influence model predictions.

These results indicate that multilingual representations 
distinguish languages while preserving similarities between related ones. Additional examples, including Spanish--German contexts,
are provided in the Appendix. This further indicates that language representations can influence one another even when they are nominally separable, which may contribute to cross-lingual interference in downstream behavior.

\subsection{Hierarchy and Simplices}
\label{results:hierarchy}
Finally, we examine whether higher-level structure emerges across languages. Across all models, we find that languages belonging to the same family organize
into  simplex-like structure in representation space, with individual
languages occupying distinct vertices.

As shown in Figure \ref{fig:germanic_simplex_qwen}, for the Germanic family, languages form a three-dimensional simplex-like geometric structure, and the
simplex as a whole is approximately orthogonal to the corresponding family-level
Germanic direction under the causal inner product
(results for Romance family languages are available in the Appendix). This pattern is consistent with
 expectations \cite{park2024geometrycategoricalhierarchicalconcepts} for hierarchical categorical concepts and indicates
that language identity and family membership are encoded as approximately
separable factors.

In addition to this hierarchical organization, we observe graded geometric
relationships within each simplex-like structure. Languages that are linguistically closer are
positioned nearer to one another, reflecting shared lexical, phonological, or
orthographic structure. Together, these results suggest that multilingual
representations combine approximate causal separability with hierarchical
organization and continuous linguistic similarity.

Taken together, these results suggest that multilingual representations are
structured in a way that supports partial predictability of cross-lingual
effects. In particular, approximate separability enables targeted diagnostics,
while structured deviations indicate when interventions in one language may
affect others. This provides a  basis for a diagnostic for anticipating and evaluating
cross-lingual behavior in deployed systems.

\section{Discussion}

As shown by the projection analyses in Fig.~\ref{fig: linear_directions_qwen} and the causal inner product heatmap in Fig.~\ref{fig:qwen_heatmap}, languages from the same family often admit stable and well-aligned concept directions even when individual languages might be comparatively underrepresented. This suggests that shared lexical, orthographic, or morphological structure can partially compensate for data scarcity, consistent with shared representational structure.

Some overlap between language concepts may arise from surface features of the languages rather than shared meaning \cite{park2024linearrepresentationhypothesisgeometry}. For example, as seen in the causal inner product  (Fig.~\ref{fig:qwen_heatmap}), concepts involving German exhibit structured association with capitalization-related feature directions. This aligns with known orthographic conventions, such as systematic noun capitalization in German and the capitalization of days and months in English, which introduce correlations between language identity and capitalization-related feature directions. These effects are consistent with how surface conventions can shape geometric representations without undermining conceptual separability.




The simplex-like  structure indicates
that LLMs encode language identity hierarchically: family membership captures
coarse-grained linguistic commonality, while positions within the simplex encode
finer-grained language-specific distinctions. The resulting geometry extends
causal–geometric analyses to multilingual settings where categories are discrete but not independent, and exhibit graded similarity. In this view, deviations
from strict orthogonality reflect meaningful hierarchical organization.

A key implication of our findings is that multilingual representations are
neither fully independent nor fully entangled. Instead, languages are largely
separable, with structured overlap that reflects linguistic similarity. As a
result, interventions applied in one language may generalize predictably to
related languages, rather than failing arbitrarily, while still exhibiting
systematic cross-lingual side effects.

This has direct implications for multilingual deployment. Structured overlap
between language representations can lead to cross-lingual transfer, bias, or
interference, even when concepts appear mostly separable. For example, a safety
filter developed in English may partially transfer to Spanish or Portuguese due
to representational similarity, while failing to transfer cleanly to more
distant languages. Our results suggest that such behavior can be anticipated
from the geometry of language representations.

From a trustworthiness perspective, this highlights the need to evaluate
multilingual systems not only within each language, but also for cross-lingual
consistency and unintended interactions. Measuring separability and similarity
jointly provides a practical way to diagnose when interventions are likely to
generalize and when cross-lingual effects should be expected.

\section{Conclusion}
Our work shows that, in multilingual representations in LLMs, language identity behaves approximately like a separable factor under causal--geometric diagnostics, with structured overlap that reflects  linguistic relationships. This is relevant because many interpretability, safety, and control methods implicitly assume that languages can be treated as independent concepts.  By providing causal–geometric diagnostics for multilingual representations, this work offers a way to assess when interventions, filters, or monitoring mechanisms designed for one language are likely to generalize safely to others, and when they may introduce cross-lingual side effects. More broadly, our findings suggest that departures from idealized geometric structure can be treated as indicators of meaningful similarity and hierarchy. Understanding and measuring this structure is  a  promising diagnostic tool for reliable multilingual deployment. 



\section{Limitations}
Our study considers a restricted multilingual setting, with concepts drawn from two Indo-European families and limited to the Latin alphabet. While this enables controlled analysis, it limits the generality of our conclusions to other scripts, low-resource languages, and non-Indo-European families. We also explore only a limited range of layers, prompts, and intervention strengths; broader sweeps may reveal additional structure. A more controlled direction would use aligned corpora (same content across languages), which we leave to future work.

Tokenization introduces unavoidable noise in counterfactual construction \cite{singh-etal-2019-bert}. Some words decompose into multiple tokens (e.g., \textit{princess} $\rightarrow$ \textit{prin} + \textit{cess}), while others appear as substrings of unrelated tokens (e.g., French \textit{bas}, \textit{est}). Although multi-token words can be filtered out, substring effects cannot be fully mitigated, implying that causally separable concepts are only \emph{approximately} orthogonal under the estimated causal inner product. 

While we briefly explored additive interventions as a diagnostic of causal signal, we found that generation-level control remains limited and highly sensitive to tokenization and context, even when geometric structure is well-defined.

\section{Future Work}

Our work points to multiple open questions for future study. Future work could analyze how causal separability and hierarchical structure evolve across layers, shedding light on when and how multilingual abstractions form during inference. Another promising  direction is whether deviations from causal separability can serve as indicators of translation uncertainty, cross-lingual interference, or limited language familiarity in multilingual models. Finally, our findings raise the question of whether causal–geometric structure could be used as a training-time diagnostic or objective, maintaining separation without discarding similarity between related concepts. 

\section*{Acknowledgements}
AS acknowledges support from the Future of Life Institute PhD Fellowship. 
This work is supported by the UKRI grant: Turing AI Fellowship EP/W002981/1. AC acknowledges support from a UKRI AI World Leading Researcher Fellowship (grant number EP/W002949/1), and from a University of Oxford UKRI Impact Acceleration Account Seed Fund award.
JY acknowledges support from Microsoft Ltd and was supported by the EPSRC grant EP/W024330/1.

\section*{Impact Statement}

This work contributes to understanding how multilingual language models represent and relate different languages. By highlighting that language representations are only approximately separable, our findings suggest that interventions, safeguards, or alignment mechanisms applied in one language may have unintended effects in others. This has implications for the safe and reliable deployment of multilingual AI systems, particularly in settings where consistent behavior across languages is critical.



\bibliographystyle{icml2026}

\newpage
\appendix


\section{Data}
We draw on three semantic relations from the Big Analogy Test Set (BATS 3.0) \cite{gladkova-etal-2016-analogy} for bilingual concepts. From the 8 selected languages, we generate 28 bilingual pairings, yielding a total of 31 sets of counterfactual word pairs. Each counterfactual pair is intended to  differ only in terms of the targeted concept.  

\section{Steering Intervention}
\label{app:steering}

Given a context representation $\lambda(x)$ and an embedding-space concept
representation $\bar{\lambda}_C$, we perform additive interventions according to
\begin{equation}
\lambda_{C,\alpha}(x) = \lambda(x) + \alpha\,\bar{\lambda}_C,
\quad \alpha \in [0,1].
\end{equation}
We evaluate the effect of these interventions by measuring log-odds
shifts and top-$k$ rank changes of target tokens associated with concept $C$. Adding $\overline{\lambda}_W$ to a context representation should increase the probability of $W$ \citep{park2024linearrepresentationhypothesisgeometry}. 
In the bilingual concept case, this can be interpreted as a diagnostic bias
toward tokens associated with the target language.

\subsection{Intervention Diagnostics}

We report additive intervention diagnostics to assess whether learned concept
directions carry causal signal at the logit level. These experiments are intended
as diagnostic probes rather than demonstrations of reliable generation control.

\begin{figure}[h]
  \centering
  \includegraphics[width=0.9\columnwidth]
    {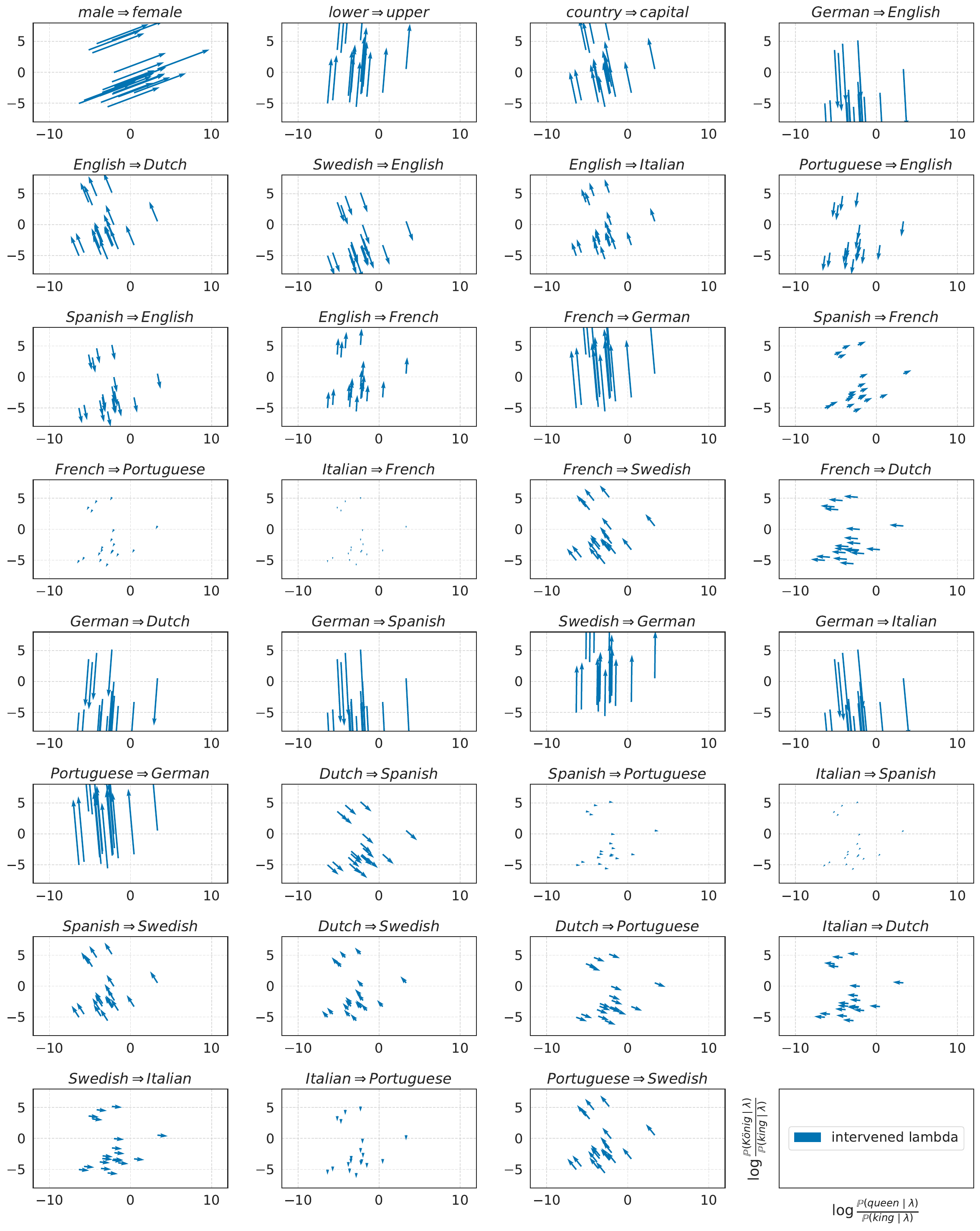}
  \caption{Additive intervention diagnostics in Qwen-3-4B  across bilingual concepts.  Each panel shows the change in
log-odds induced by the intervention, with arrows indicating how probability
mass shifts between counterfactual token pairs as a function of intervention
strength. Across
concepts, interventions induce systematic but heterogeneous logit-level shifts,
illustrating that while causal signal is present, steering effects are limited
and vary with language pair and tokenization.}
\label{steering_fig_qwen}
\end{figure}

\begin{figure}[h]
  \centering
  \includegraphics[width=0.9\columnwidth]
    {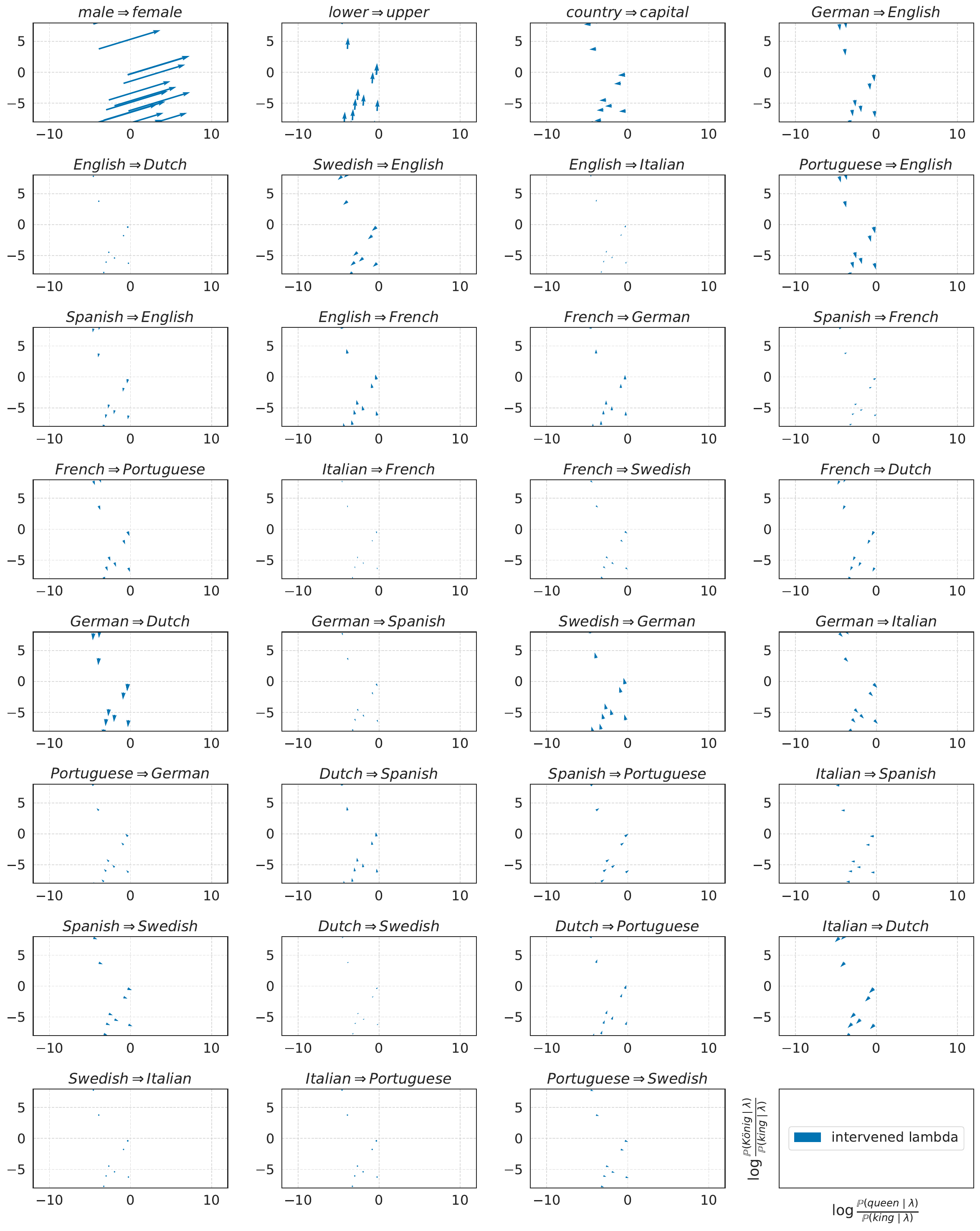}
  \caption{Additive intervention diagnostics in Mistral-7B-v0.3.}
  \label{steering_fig_mistral}
\end{figure}

\begin{figure}[h]
  \centering
  \includegraphics[width=0.9\columnwidth]{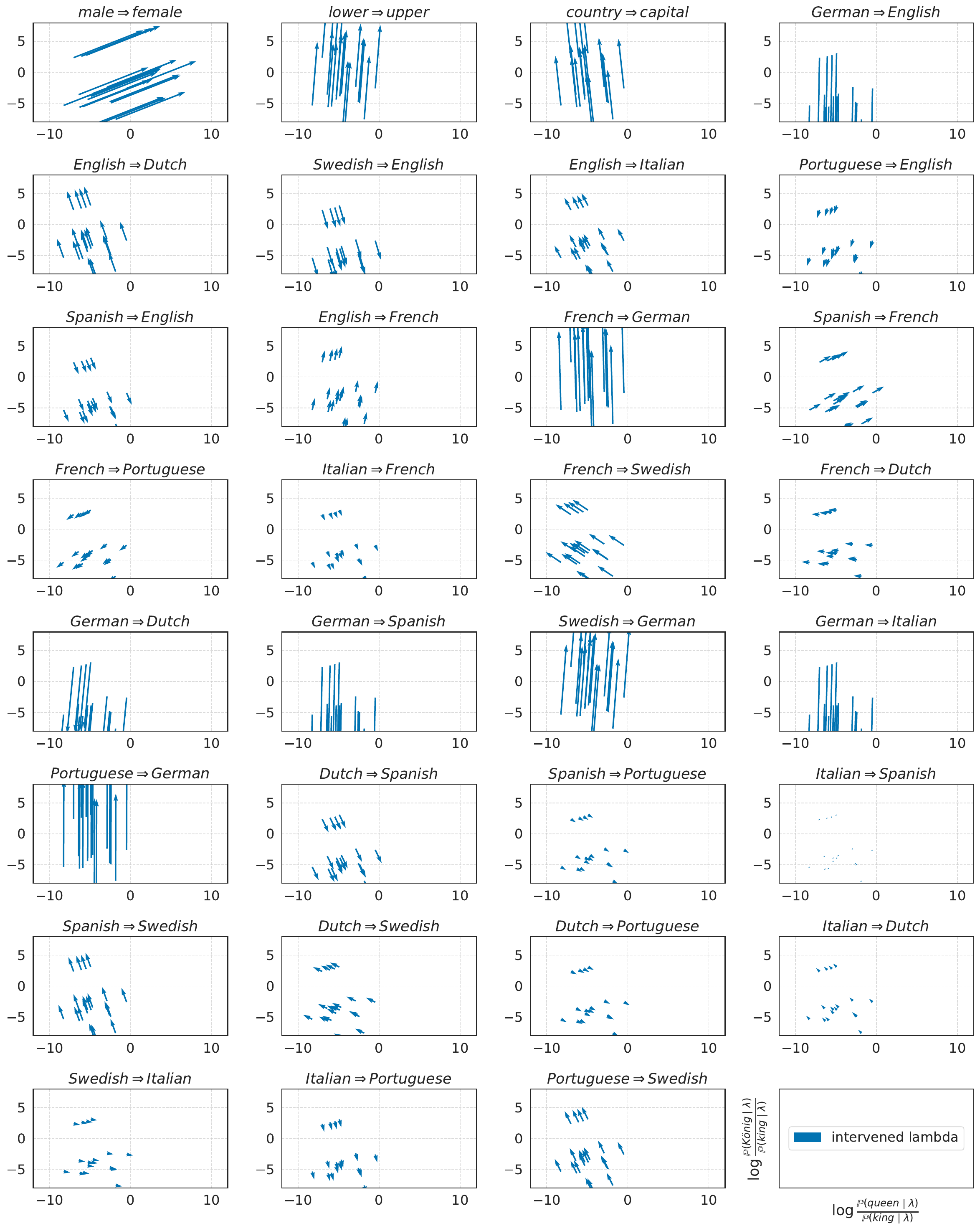}
  \caption{Additive intervention diagnostics in LLama-3-8B.}
\label{steering_fig_llama}
\end{figure}

We apply additive interventions along estimated concept directions
to the model’s internal representations and measure the resulting changes in
next-token log-odds for target counterfactual tokens.  As shown in Figures \ref{steering_fig_qwen}, \ref{steering_fig_mistral} and \ref{steering_fig_llama} across both simple categorical concepts and bilingual  contrasts, we
observe systematic and approximately monotonic shifts in log-odds in the
expected direction as the intervention strength increases.  This indicates that
the learned directions are not merely descriptive probes, but are associated with intervention-relevant structure in the model's representations. However, these diagnostics further suggest  that, despite representation-level signal relevant to intervention, reliable generation-level control would require addressing tokenization and autoregressive effects.


\section{Additional Figures}
\label{app:additional_figures}


\begin{figure*}[p]
  \centering
  \includegraphics[width=\textwidth]{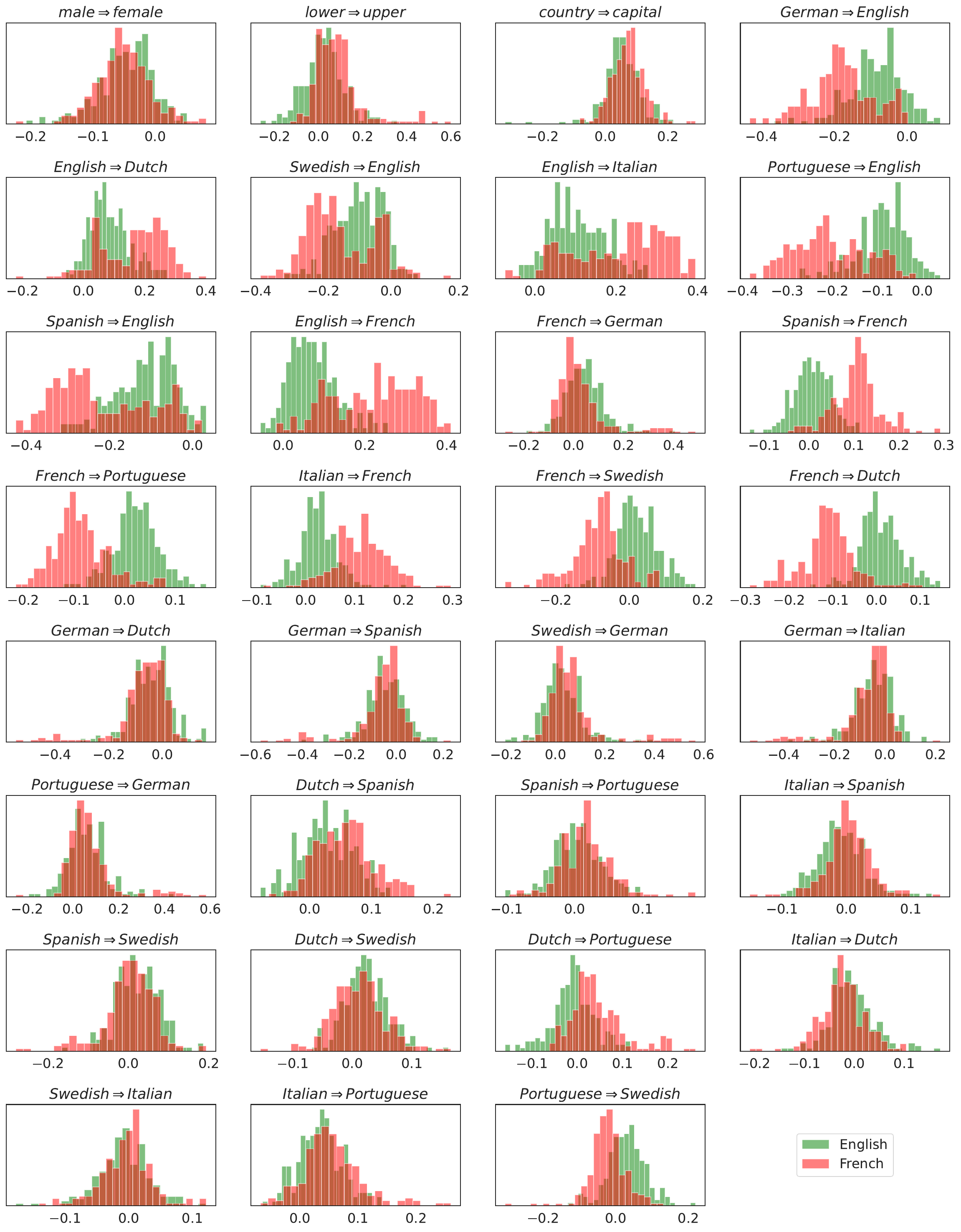}
  \caption{Histogram of $\mathbf{\gamma}_C^\top \lambda(x_j^{\text{en}})$ vs $\mathbf{\gamma}_C^\top \lambda(x_j^{\text{fr}})$ for  concepts $C$  for Qwen3-4B, where $\{ x_j^{\text{en}} \}$ are random contexts from English Wikipedia, and $\{ x_j^{\text{fr}} \}$ are random contexts from French Wikipedia.}
  \label{fig:qwen-measure-en-fr}
\end{figure*}

\clearpage
\begin{figure*}[p]
  \centering
  \includegraphics[width=\textwidth]{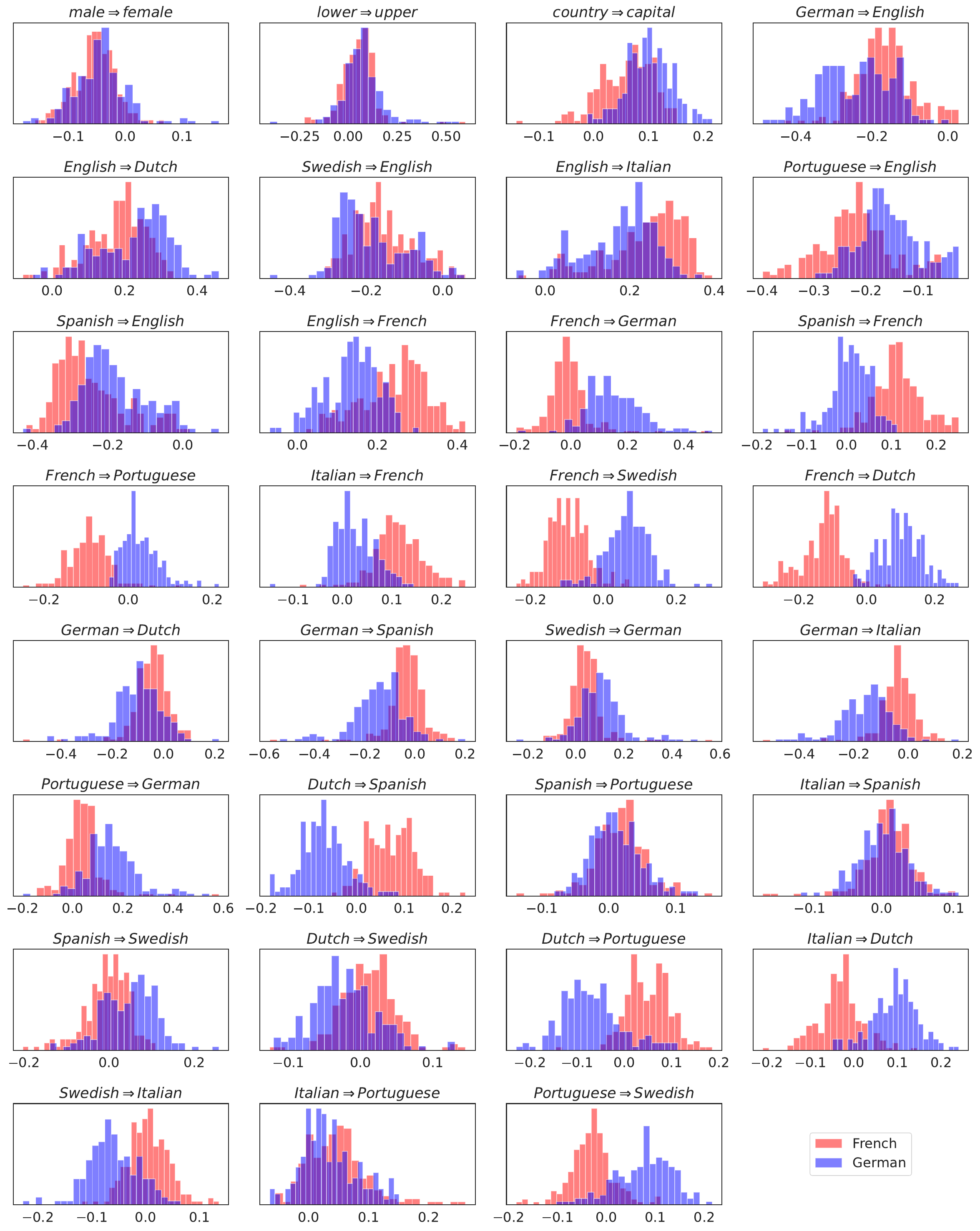}
  \caption{Histogram of $\mathbf{\gamma}_C^\top \lambda(x_j^{\text{fr}})$ vs $\mathbf{\gamma}_C^\top \lambda(x_j^{\text{de}})$ for  concepts $C$  for Qwen3-4B, where $\{ x_j^{\text{fr}} \}$ are random contexts from French Wikipedia, and $\{ x_j^{\text{de}} \}$ are random contexts from German Wikipedia. }
  \label{fig:qwen-measure-fr-de}
\end{figure*}


\clearpage
\begin{figure*}[p]
  \centering
  \includegraphics[width=\textwidth]{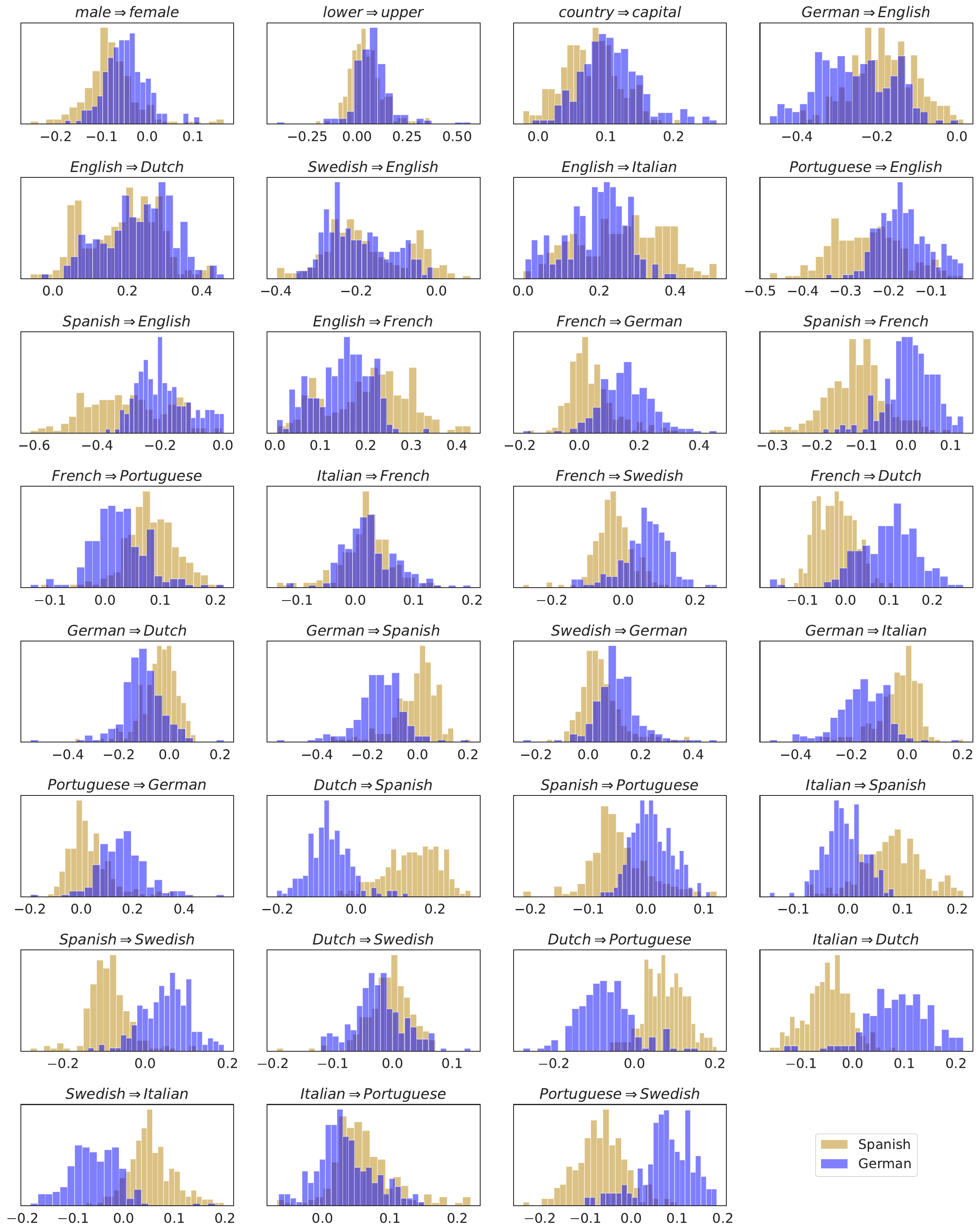}
  \caption{Histogram of $\mathbf{\gamma}_C^\top \lambda(x_j^{\text{es}})$ vs $\mathbf{\gamma}_C^\top \lambda(x_j^{\text{de}})$ for  concepts $C$  for Qwen3-4B, where $\{ x_j^{\text{es}} \}$ are random contexts from Spanish Wikipedia, and $\{ x_j^{\text{de}} \}$ are random contexts from German Wikipedia. }
  \label{fig:qwen-measure-es-de}
\end{figure*}

  




\clearpage
\begin{figure*}[p]
  \centering
  \includegraphics[width=\textwidth]{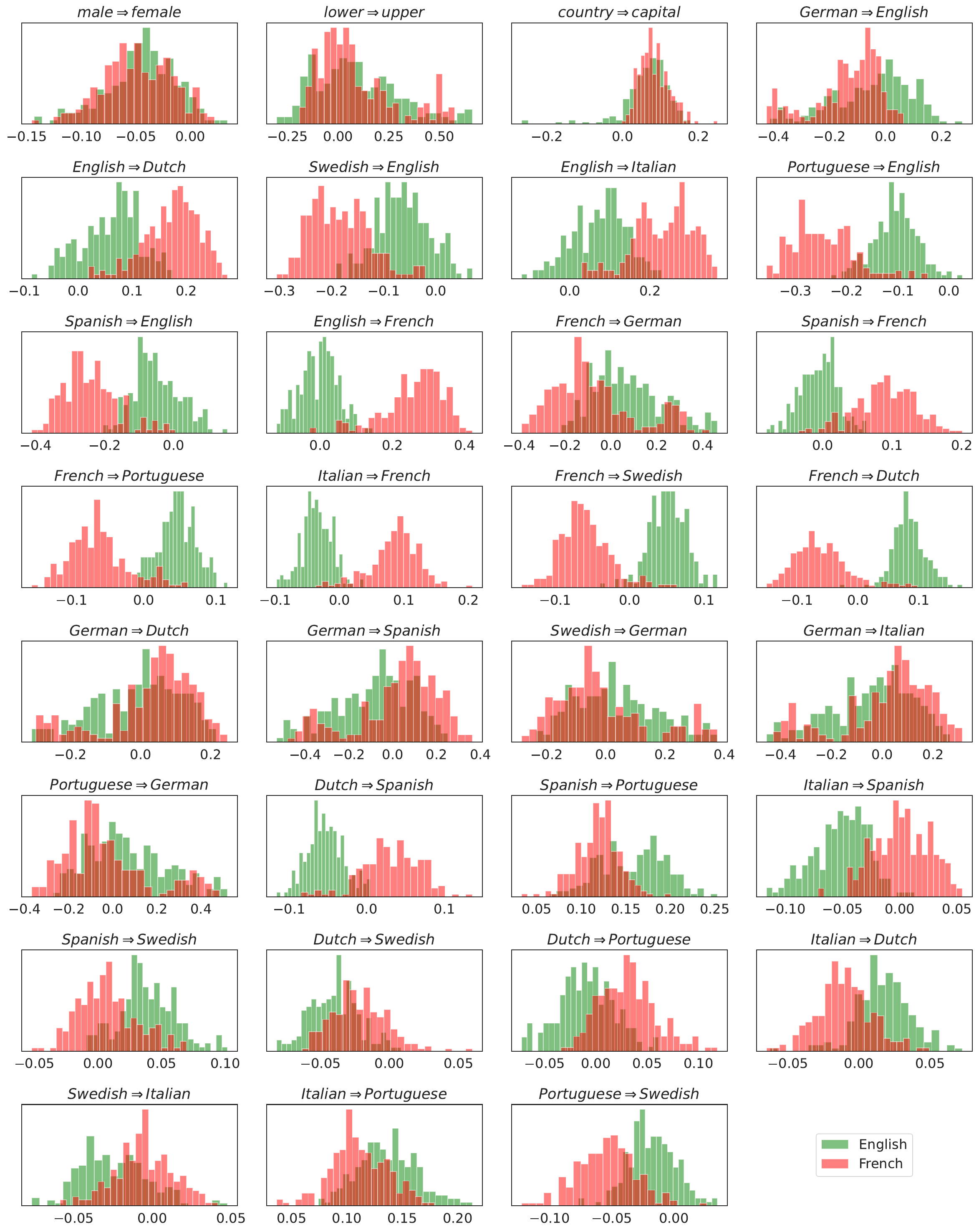}
  \caption{Histogram of $\mathbf{\gamma}_C^\top \lambda(x_j^{\text{en}})$ vs $\mathbf{\gamma}_C^\top \lambda(x_j^{\text{fr}})$ for  concepts $C$  for Mistral-7B-v0.3, where $\{ x_j^{\text{en}} \}$ are random contexts from English Wikipedia, and $\{ x_j^{\text{fr}} \}$ are random contexts from French Wikipedia. }
  \label{fig:mistral-measure-en-fr}
\end{figure*}

\clearpage
\begin{figure*}[p]
  \centering
  \includegraphics[width=\textwidth]{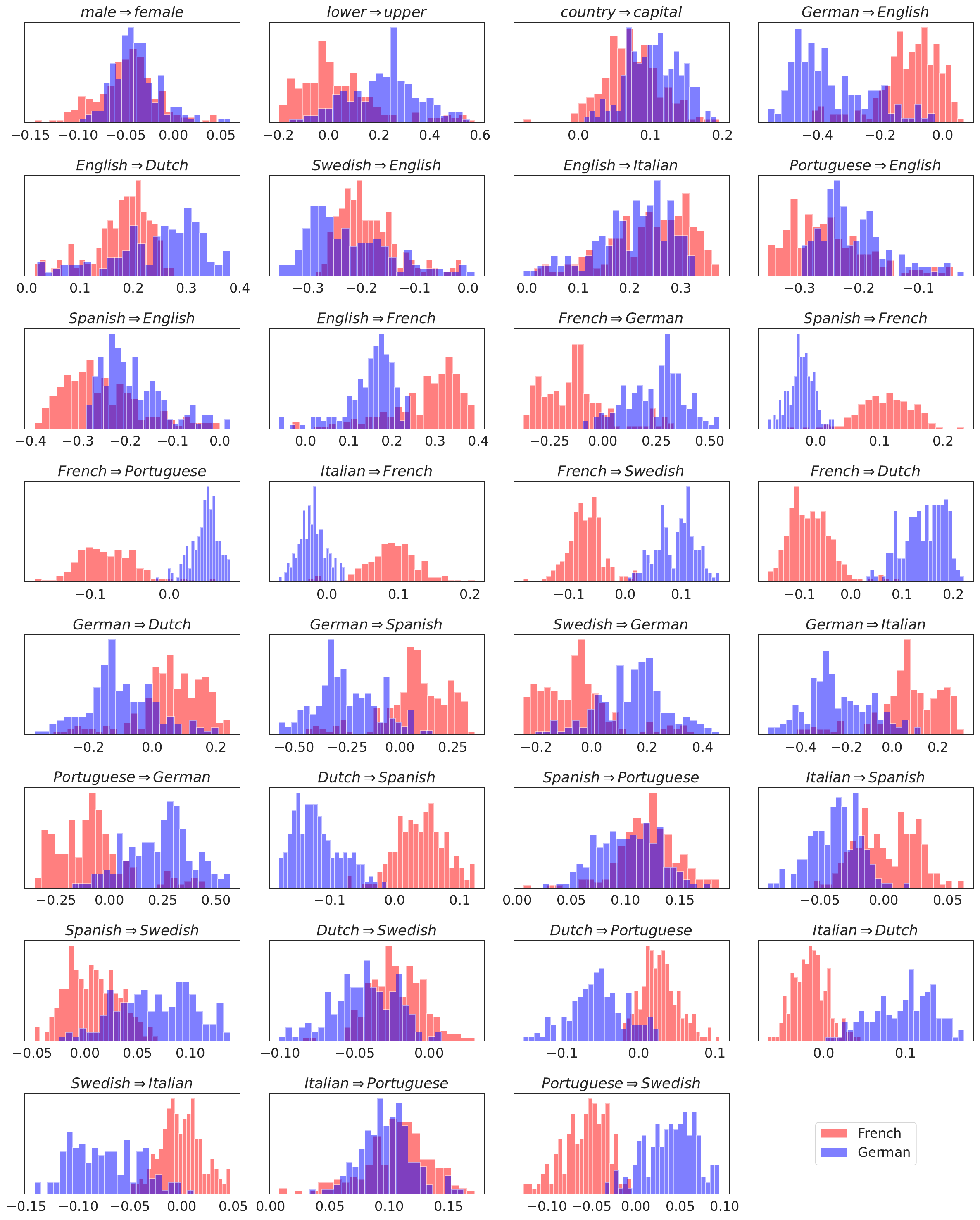}
  \caption{Histogram of $\mathbf{\gamma}_C^\top \lambda(x_j^{\text{fr}})$ vs $\mathbf{\gamma}_C^\top \lambda(x_j^{\text{de}})$ for  concepts $C$  for Mistral-7B-v0.3, where $\{ x_j^{\text{fr}} \}$ are random contexts from French Wikipedia, and $\{ x_j^{\text{de}} \}$ are random contexts from German Wikipedia. }
  \label{fig:mistral-measure-fr-de}
\end{figure*}

\clearpage
\begin{figure*}[p]
  \centering
  \includegraphics[width=\textwidth]{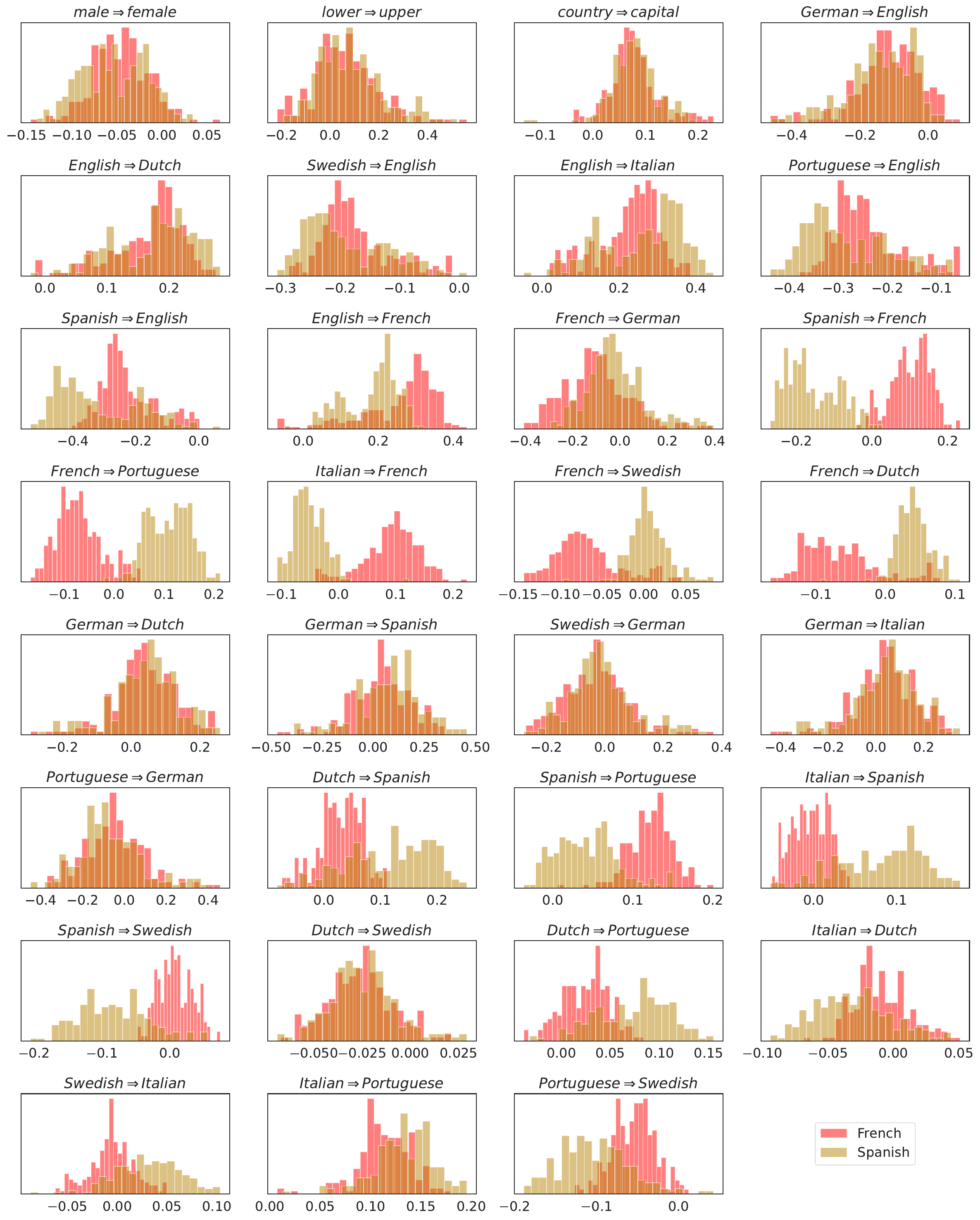}
  \caption{Histogram of $\mathbf{\gamma}_C^\top \lambda(x_j^{\text{fr}})$ vs $\mathbf{\gamma}_C^\top \lambda(x_j^{\text{es}})$ for  concepts $C$  for Mistral-7B-v0.3, where $\{ x_j^{\text{fr}} \}$ are random contexts from French Wikipedia, and $\{ x_j^{\text{es}} \}$ are random contexts from Spanish Wikipedia. }
  \label{fig:mistral-measure-fr-es}
\end{figure*}

\clearpage
\begin{figure*}[p]
  \centering
  \includegraphics[width=\textwidth]{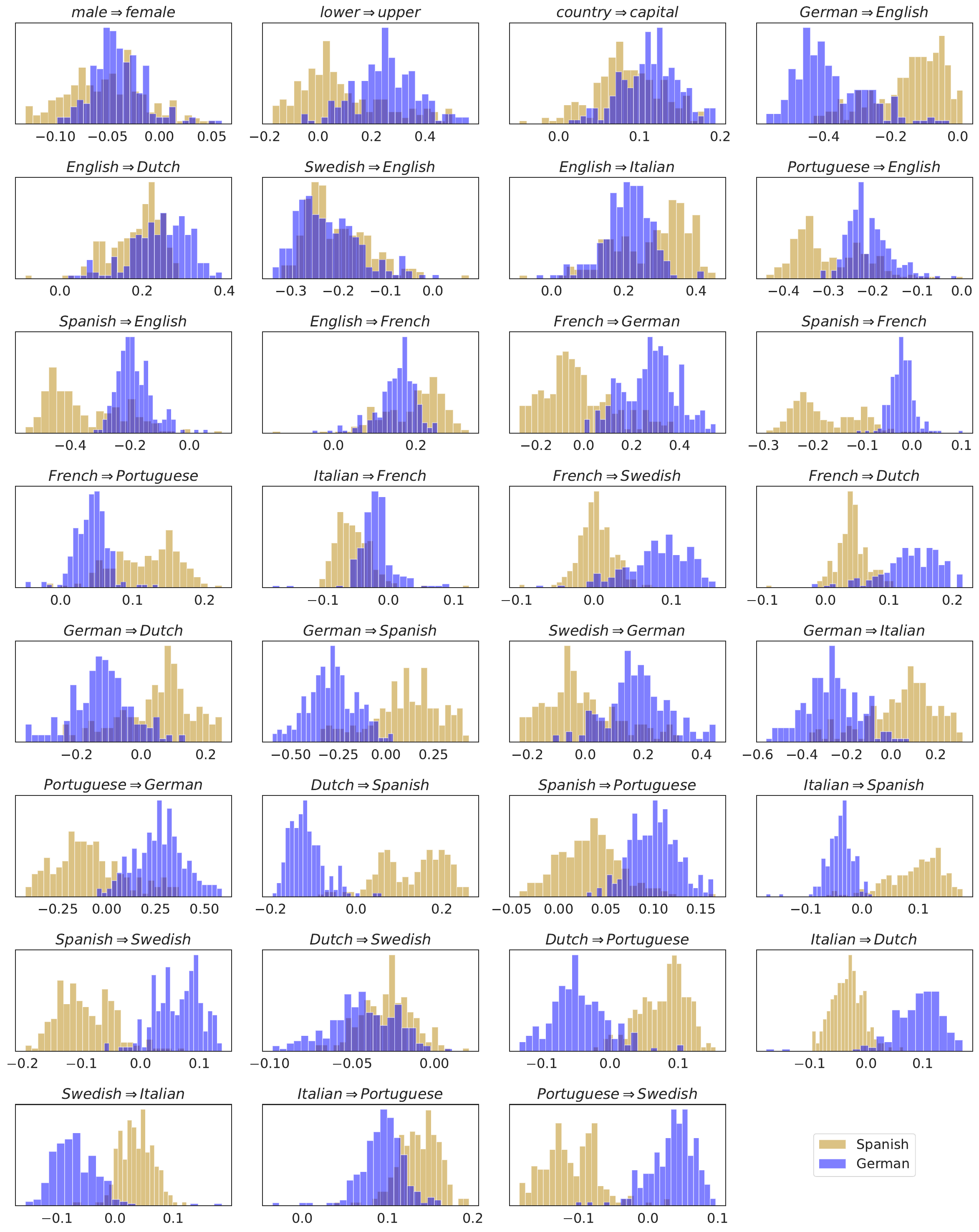}
  \caption{Histogram of $\mathbf{\gamma}_C^\top \lambda(x_j^{\text{es}})$ vs $\mathbf{\gamma}_C^\top \lambda(x_j^{\text{de}})$ for  concepts $C$  for Mistral-7B-v0.3, where $\{ x_j^{\text{es}} \}$ are random contexts from Spanish Wikipedia, and $\{ x_j^{\text{de}} \}$ are random contexts from German Wikipedia. }
  \label{fig:mistral-measure-es-de}
\end{figure*}

\begin{figure}[h]
  \centering
  \includegraphics[width=0.9\columnwidth]{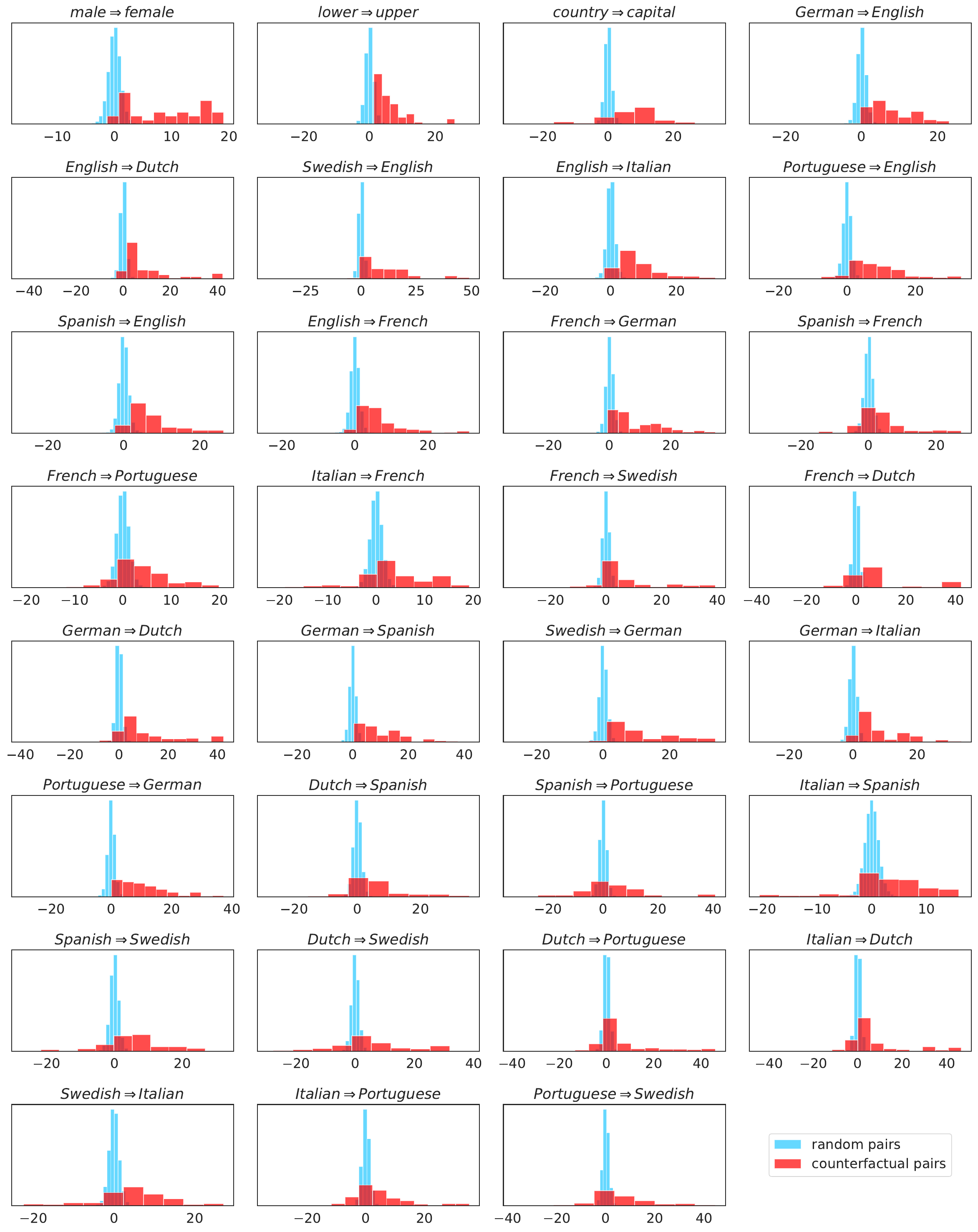}
  \caption{Projection geometry in Mistral-7B-v0.3.}
\end{figure}


\begin{figure}[h]
  \centering
  \includegraphics[width=0.9\columnwidth]{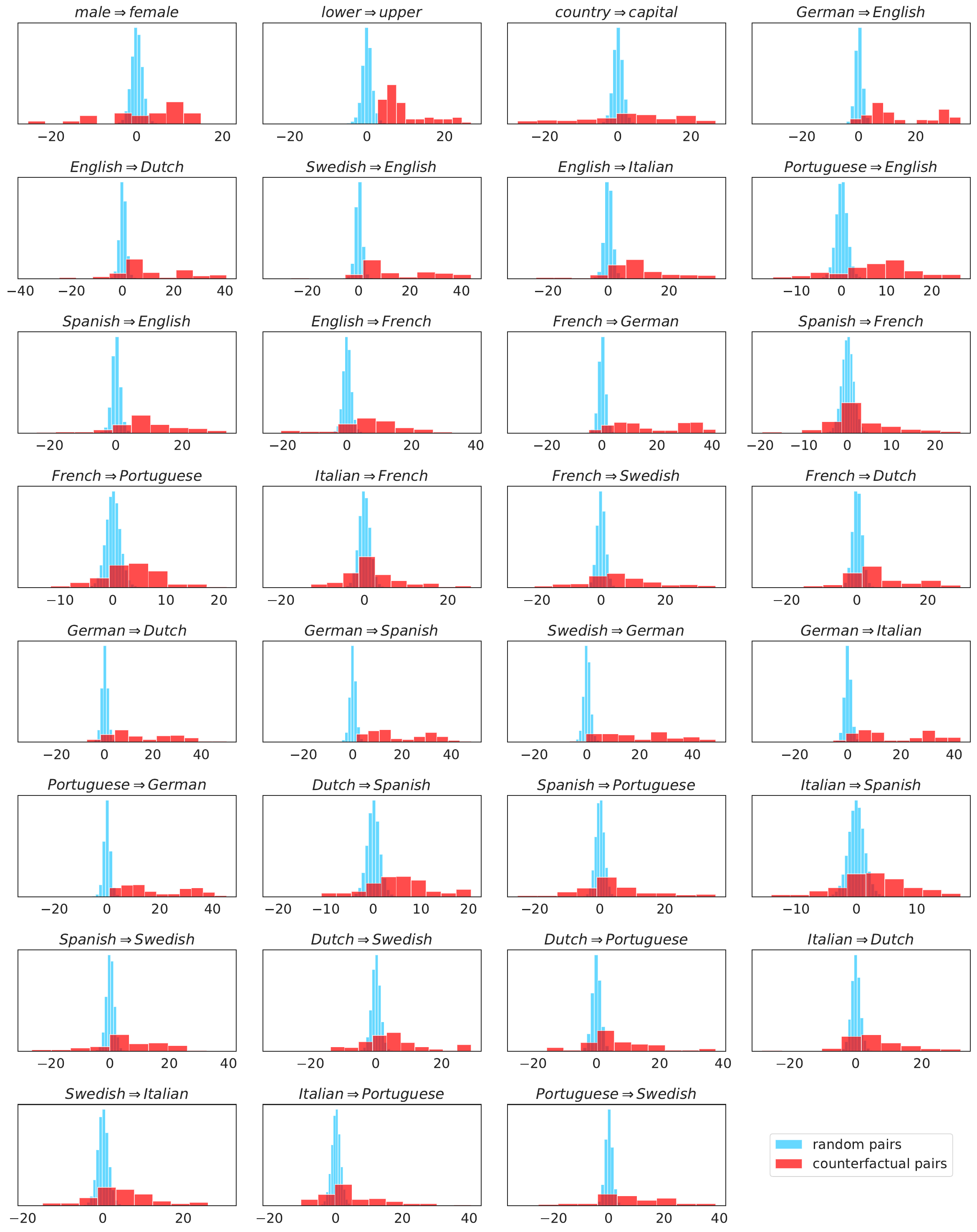}
  \caption{Projection geometry in LLama-3-8B.}
\end{figure}


\begin{figure}[h]
  \centering
  \includegraphics[width=0.9\columnwidth]{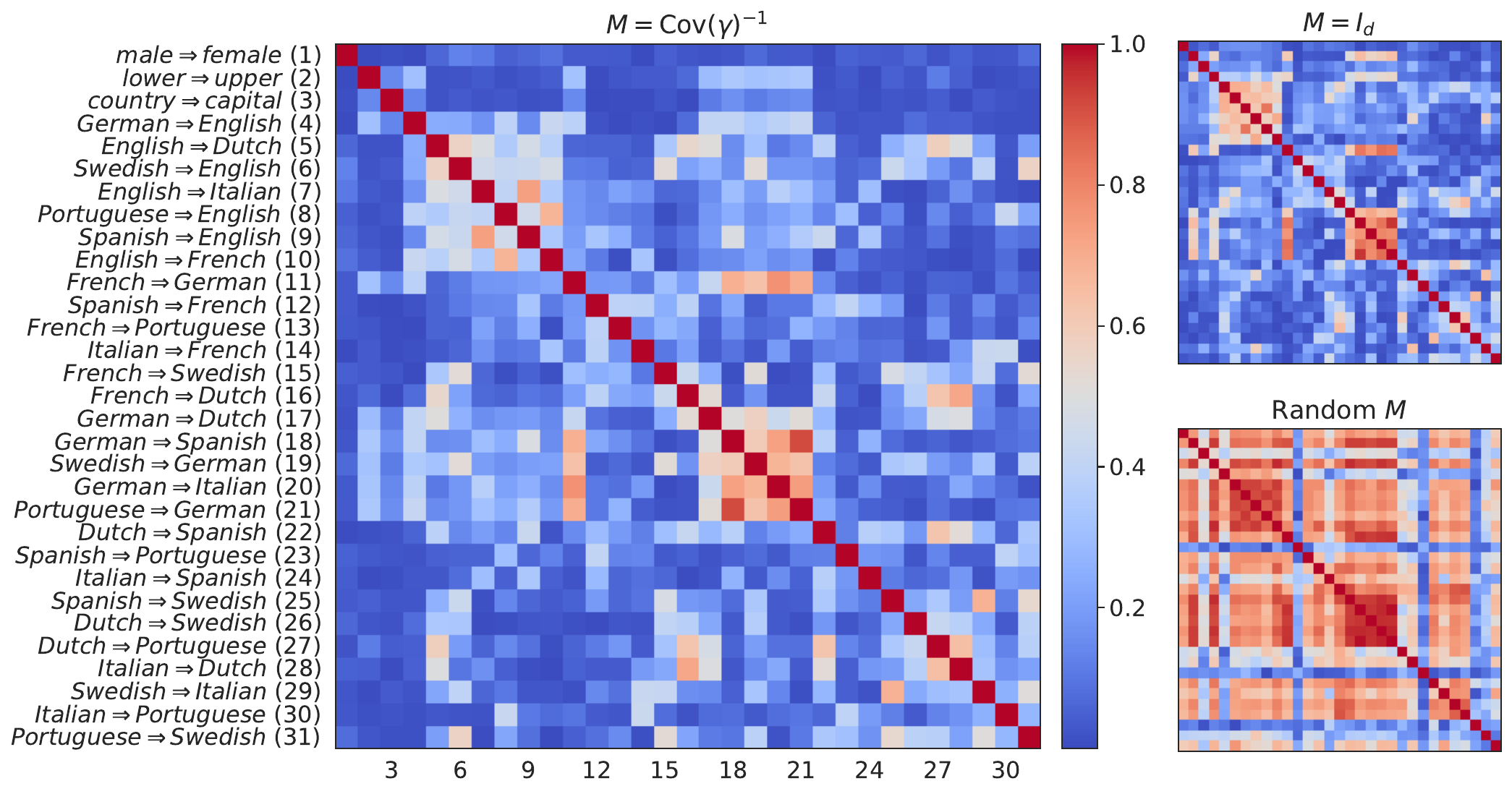}
  \caption{Causal inner product  in Mistral-7B-v0.3.}
\end{figure}


\begin{figure}[h]
  \centering
  \includegraphics[width=0.9\columnwidth]{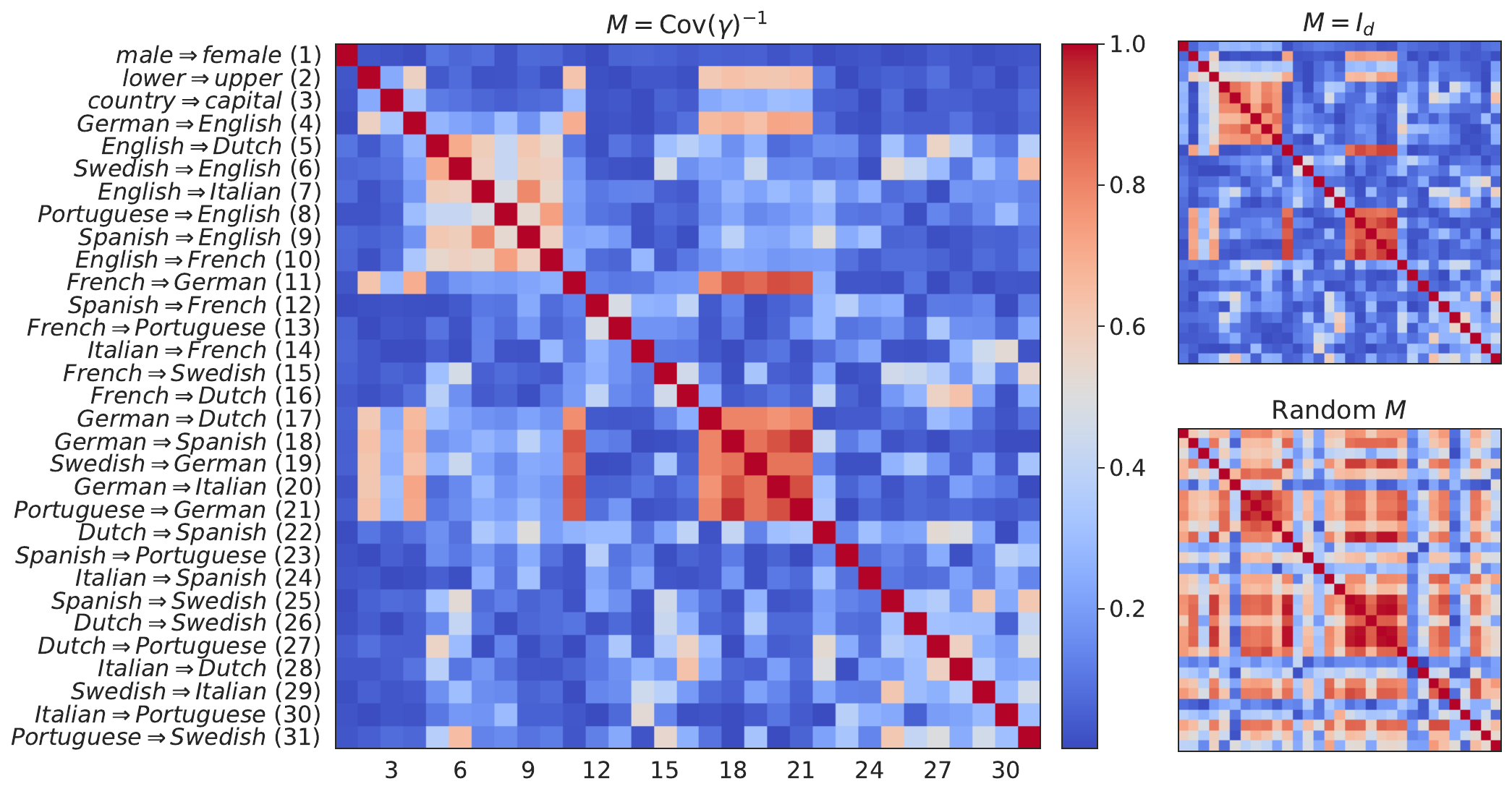}
  \caption{Causal inner product  in LLama-3-8B.}
\end{figure}




\subsection{2D and 3D Projections}
\begin{figure*}[t]
  \centering
  \includegraphics[width=\textwidth]{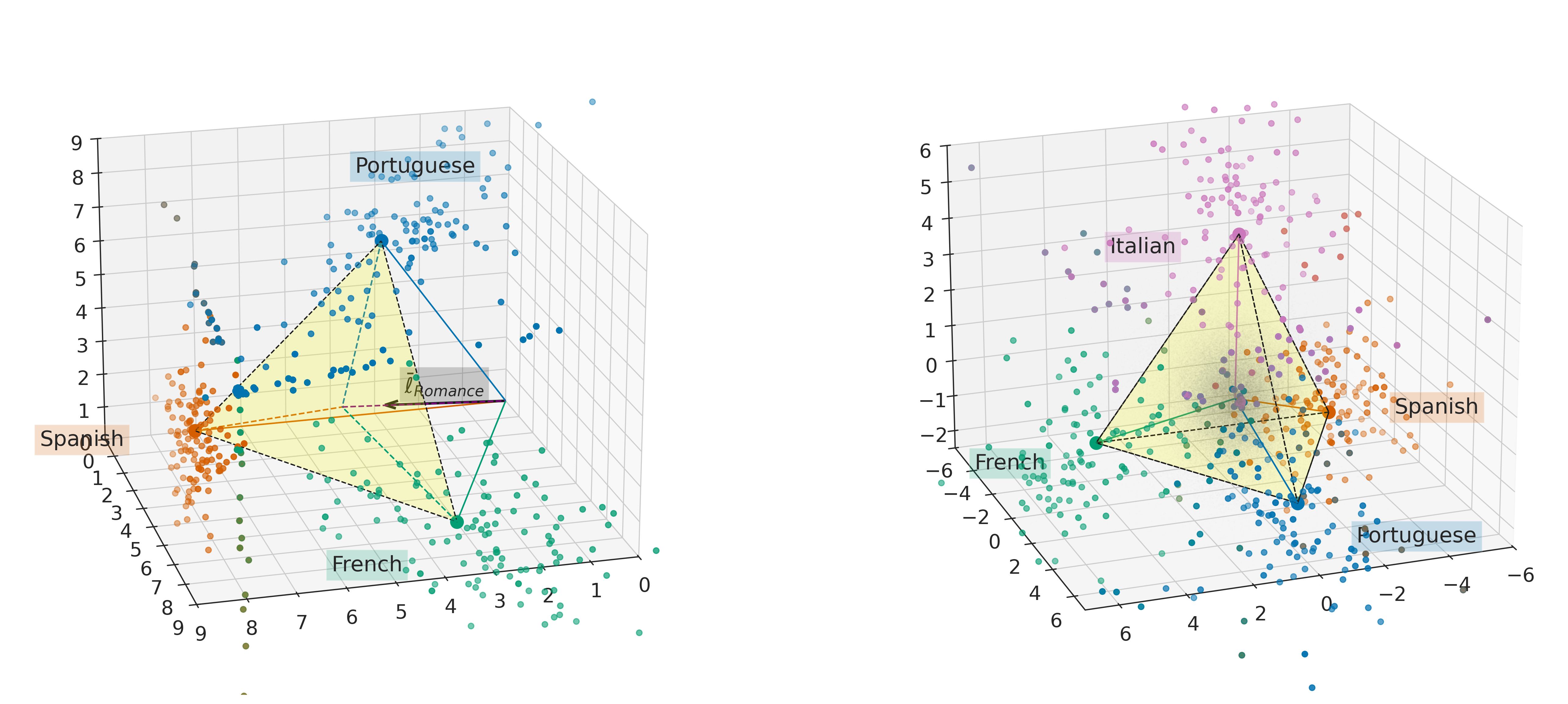}
  \caption{Qwen3-4B: Two-dimensional projections of Romance language
representations under the causal inner product.}
  \label{apfig:romance_simplex_qwen}
\end{figure*}

\begin{figure*}[t]
  \centering
  \includegraphics[width=\textwidth]{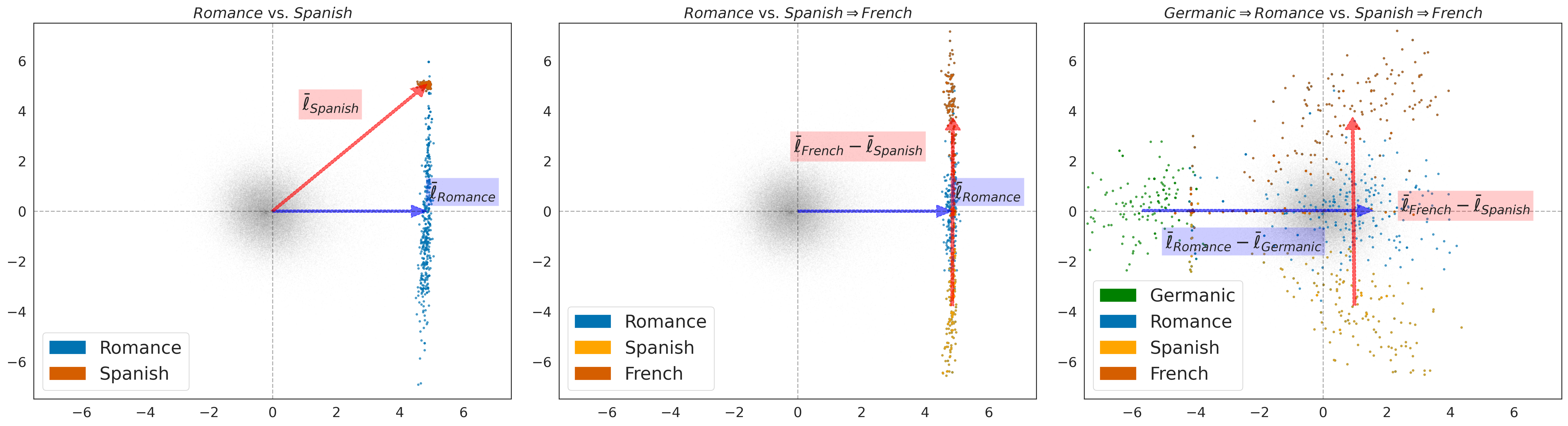}
  \caption{Qwen3-4B: Two-dimensional projection of Romance language representations under
the causal inner product. Language-specific clusters are visible, with relative
distances reflecting graded linguistic similarity within the family. As in the
Germanic case, the projection is consistent with a higher-dimensional simplex-like 
structure shown in the main paper.}
  \label{apfig:romance_simplex_qwen2}
\end{figure*}

\begin{figure*}[t]
  \centering
  \includegraphics[width=\textwidth]{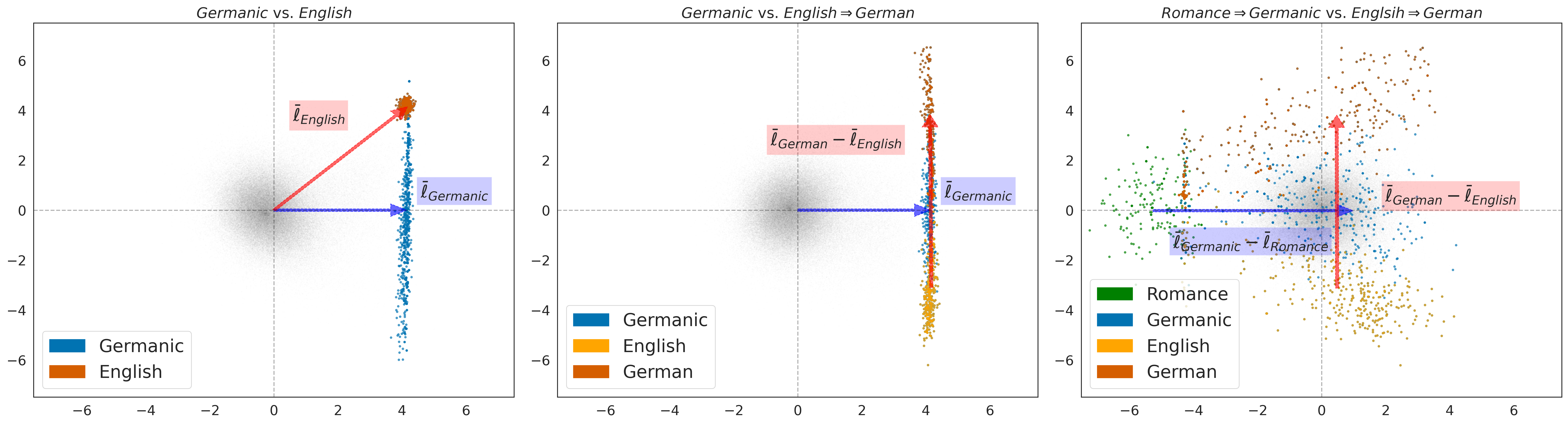}
  \caption{Qwen3-4B: Two-dimensional projection of Germanic language representations under
the causal inner product. Each cluster corresponds to tokens associated with a
specific language, while the gray background points represent randomly sampled
vocabulary tokens. The projection illustrates separation between individual
languages and approximate orthogonality between the Germanic family direction
and non-family variation, consistent with the simplex-like structure shown in
Figure~\ref{fig:germanic_simplex_qwen}}
  \label{apfig:romance_simplex_qwen3}
\end{figure*}

\begin{figure*}[t]
  \centering
  \includegraphics[width=\textwidth]{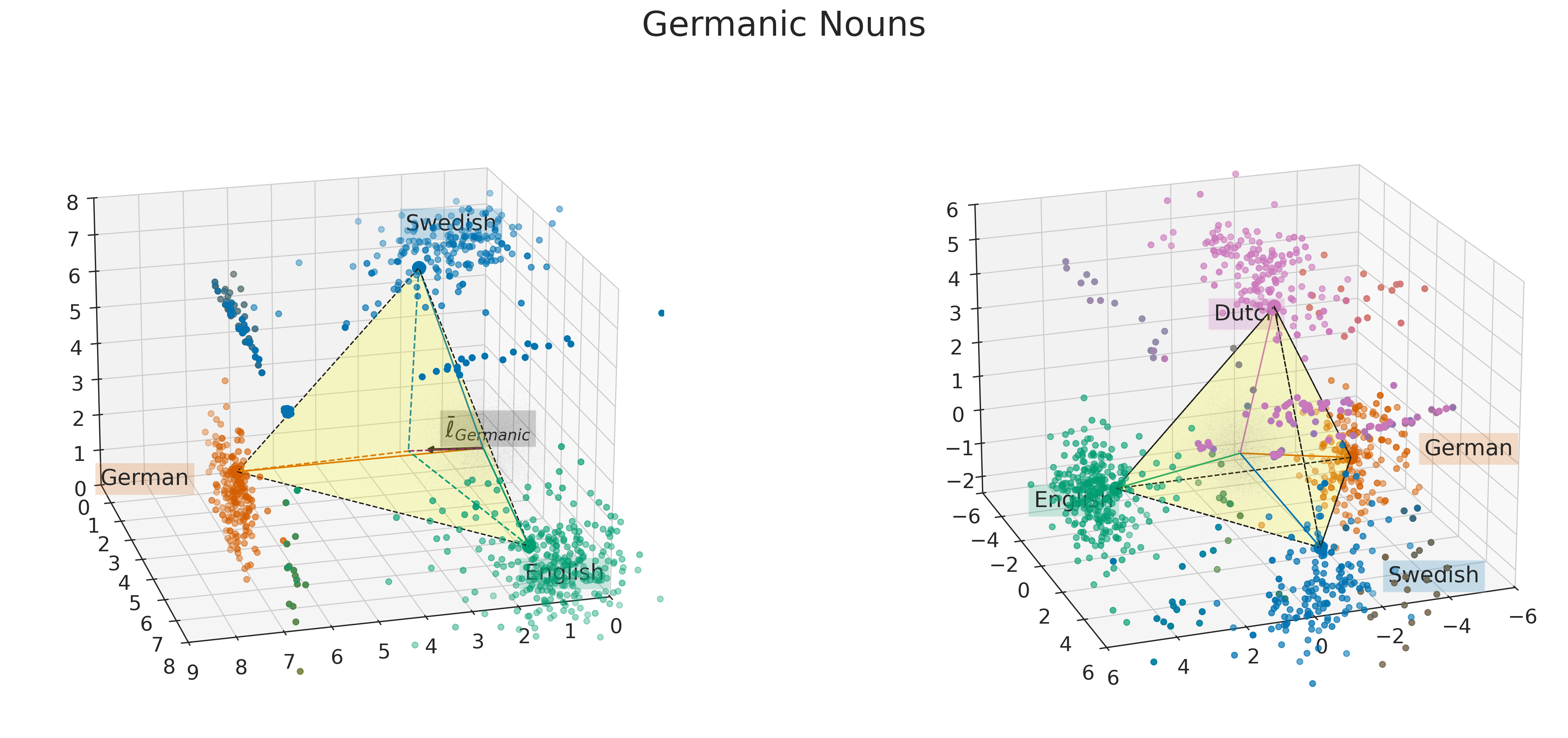}
  \caption{Mistral-7B-v0.3: Two-dimensional projections of Germanic  language
representations under the causal inner product.}
  \label{apfig:germanic_simplex_mistral}
\end{figure*}

\begin{figure*}[t]
  \centering
  \includegraphics[width=\textwidth]{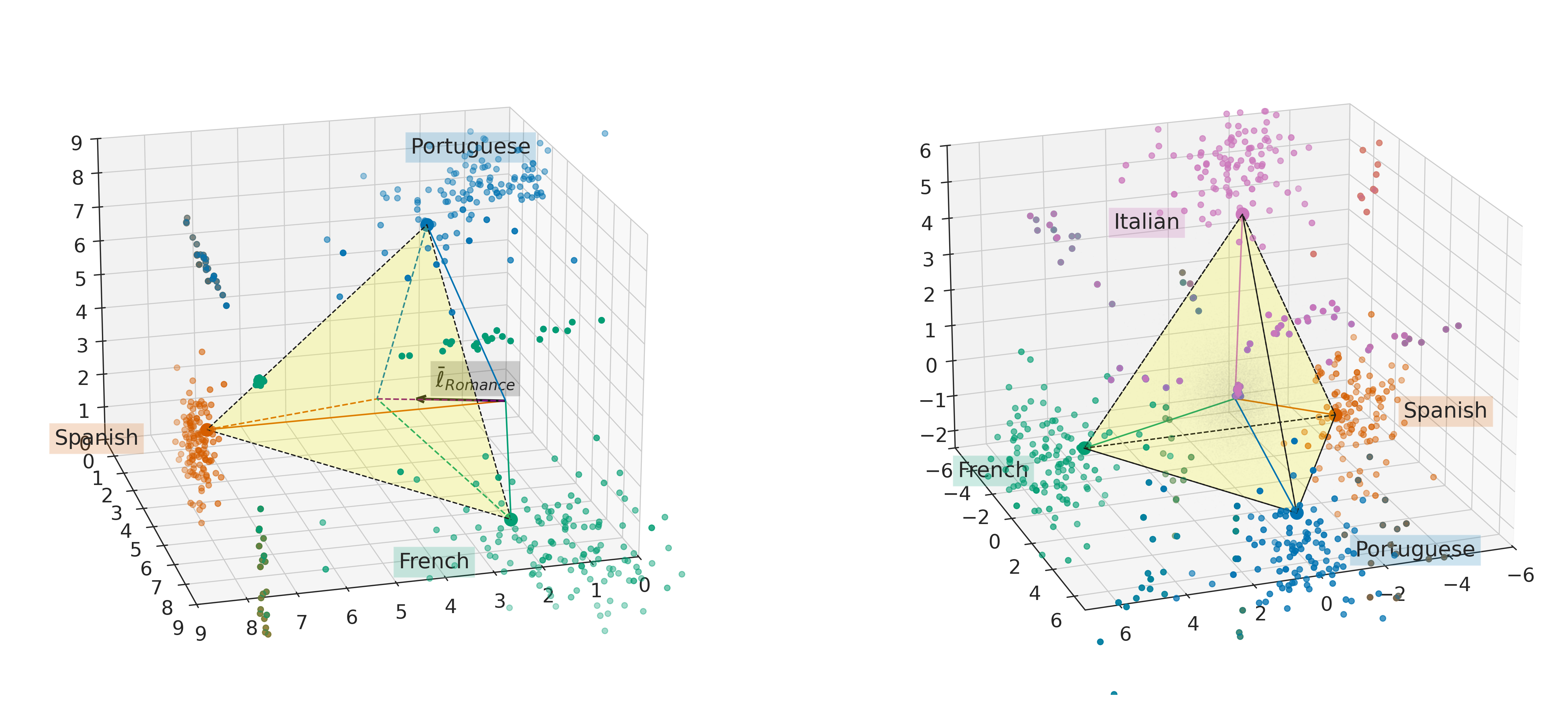}
  \caption{Mistral-7B-v0.3: Two-dimensional projections of Romance language
representations under the causal inner product.}
  \label{apfig:romance_simplex_mistral2}
\end{figure*}

\begin{figure*}[t]
  \centering
  \includegraphics[width=\textwidth]{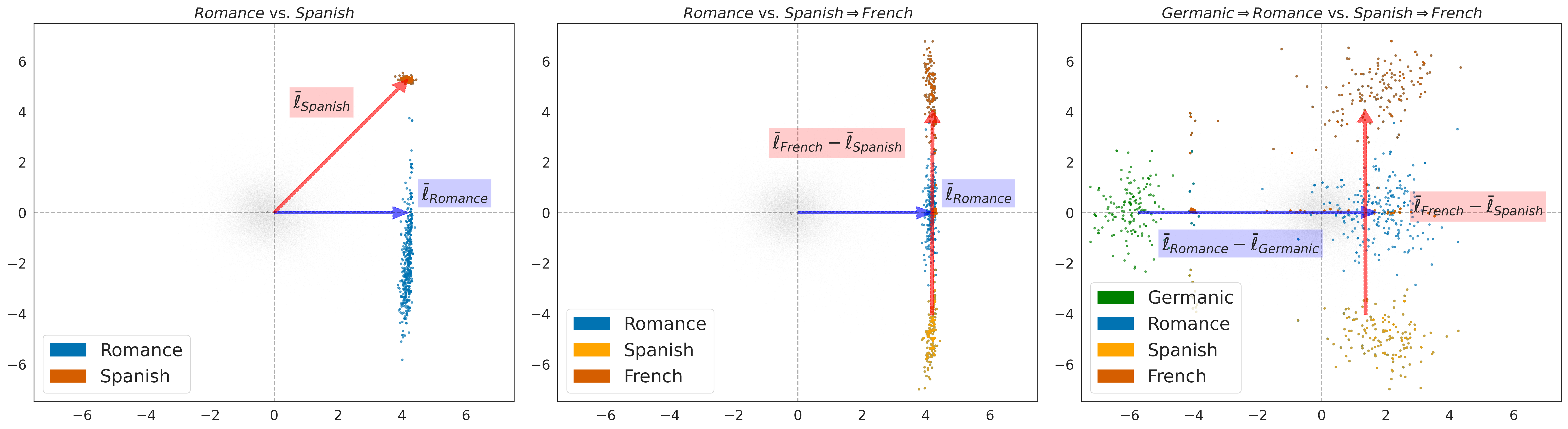}
  \caption{Mistral-7B-v0.3: Two-dimensional projection of Romance language representations under
the causal inner product. }
  \label{apfig:romance_simplex_mistral3}
\end{figure*}

\begin{figure*}[t]
  \centering
  \includegraphics[width=\textwidth]{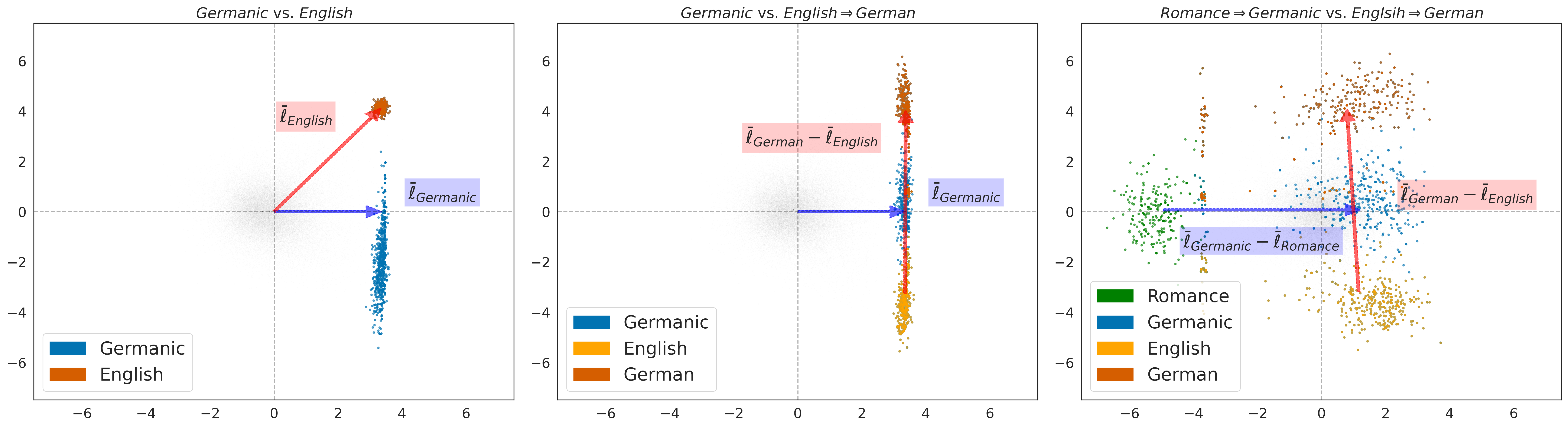}
  \caption{Mistral-7B-v0.3: Two-dimensional projection of Germanic language representations under
the causal inner product.}
  \label{apfig:germanic_simplex_mistral4}
\end{figure*}

\begin{figure*}[t]
  \centering
  \includegraphics[width=\textwidth]{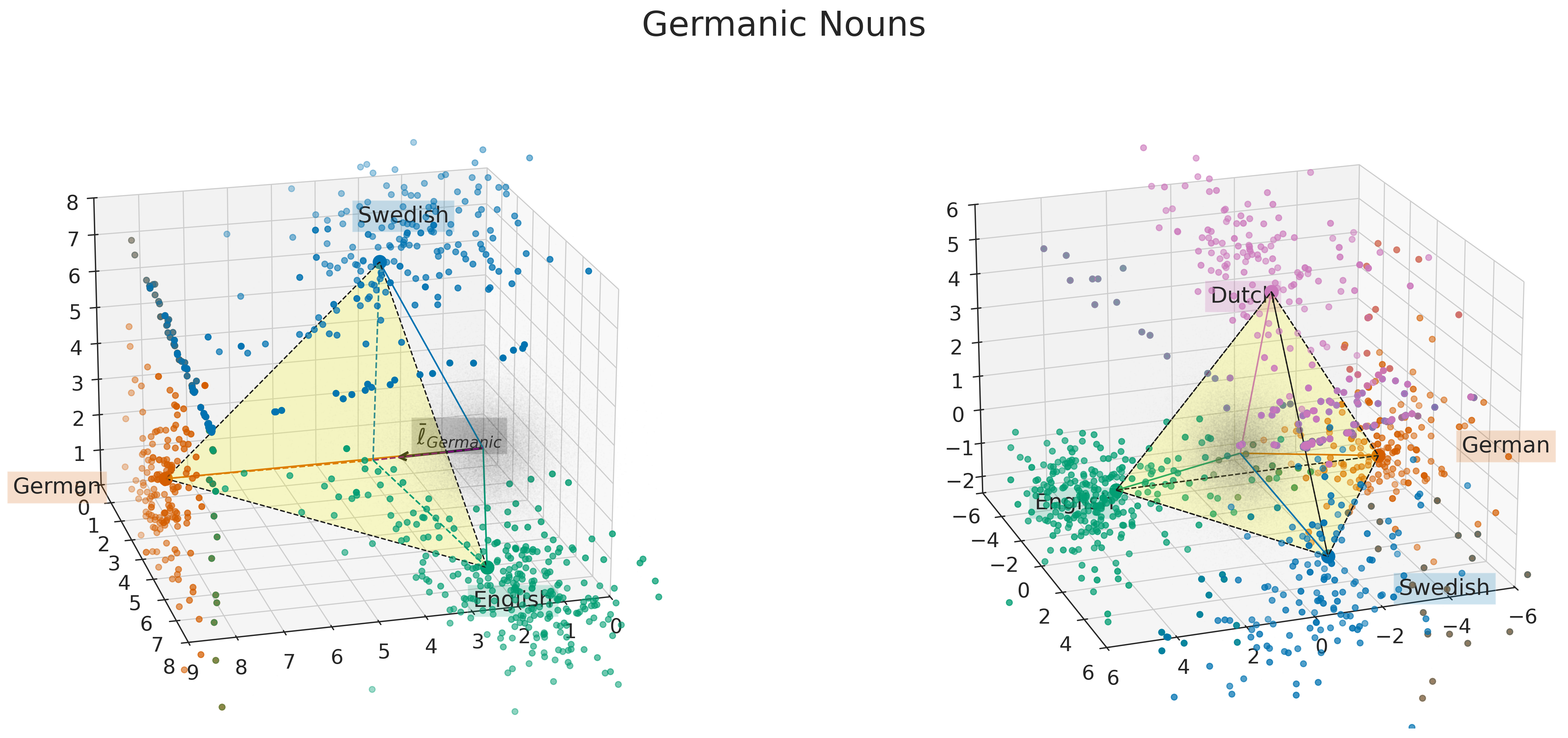}
  \caption{Llama-3-8B: Two-dimensional projections of Germanic  language
representations under the causal inner product.}
  \label{apfig:germanic_simplex_llama}
\end{figure*}

\begin{figure*}[t]
  \centering
  \includegraphics[width=\textwidth]{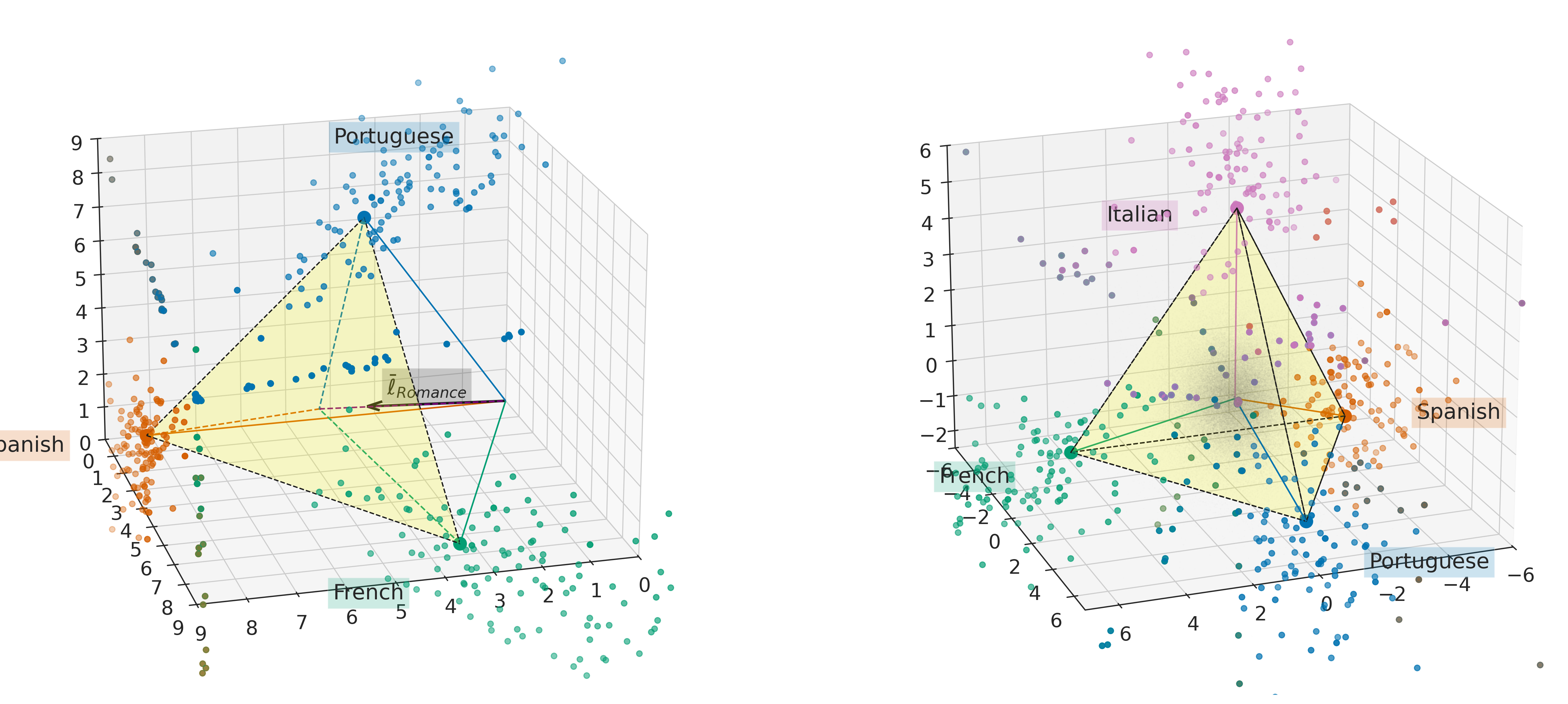}
  \caption{Llama-3-8B: Two-dimensional projections of Romance language
representations under the causal inner product.}
  \label{apfig:romance_simplex_llama2}
\end{figure*}

\begin{figure*}[t]
  \centering
  \includegraphics[width=\textwidth]{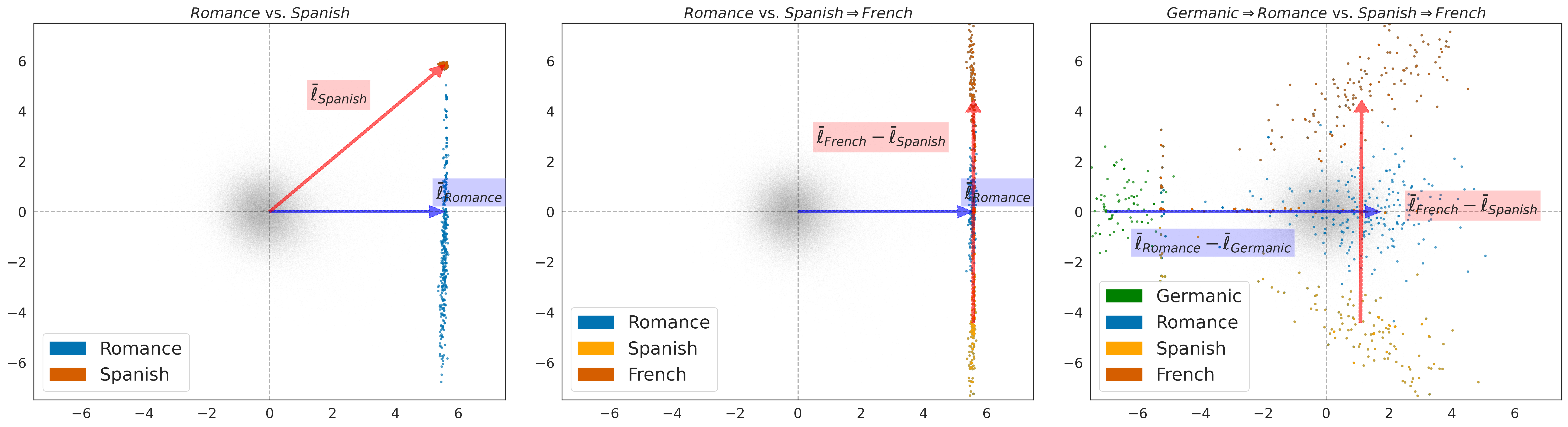}
  \caption{Llama-3-8B: Two-dimensional projection of Romance language representations under
the causal inner product. }
  \label{apfig:romance_simplex_llama3}
\end{figure*}

\begin{figure*}[t]
  \centering
  \includegraphics[width=\textwidth]{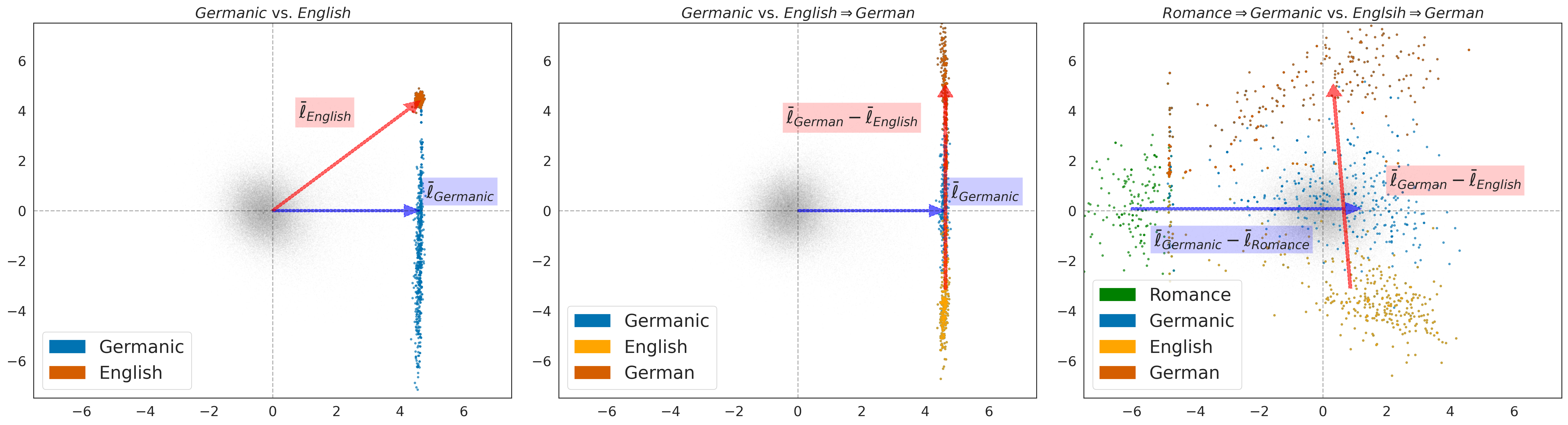}
  \caption{Llama-3-8B: Two-dimensional projection of Germanic language representations under
the causal inner product.}
  \label{apfig:germanic_simplex_llama4}

\end{figure*}

In Figures \ref{apfig:romance_simplex_qwen}-\ref{apfig:germanic_simplex_llama4} we provide additional two-dimensional and three-dimensional  projections  of language representations
under the causal inner product to complement the three-dimensional simplex-like 
visualization shown in the main paper (Fig. \ref{fig:germanic_simplex_qwen}). These plots illustrate the same
hierarchical structure in a lower-dimensional subspace and serve as supporting
evidence for the simplex geometry observed for Germanic languages.

In the three-dimensional projections, we observe that three linguistic concepts
form a $(k\!-\!1)$-dimensional simplex-like structure, corresponding to a triangle embedded in a
plane, while four concepts form a three-dimensional simplex (a tetrahedron).
In both cases, the simplex is approximately orthogonal to the corresponding
language-family direction (e.g., $\bar{\ell}_{\text{Germanic}}$), consistent
with theoretical predictions for categorical hierarchies.

Beyond hierarchy, the geometry also reflects graded linguistic similarity.
Languages that are lexically or orthographically closer are positioned nearer
to one another within the simplex, even when they are causally separable. This
extends prior observations for simpler categorical concepts (e.g., animals or
plants) to multilingual settings, where distance encodes similarity in addition
to categorical structure \cite{park2024geometrycategoricalhierarchicalconcepts}.

\end{document}